\documentclass{article}

\usepackage{PRIMEarxiv}
\usepackage{bm}
\usepackage[flushleft]{threeparttable}
\usepackage{multirow}

\usepackage{tabularx}
\usepackage{float}
\usepackage{subfig}
\usepackage{amsmath}
\usepackage{makecell}
\usepackage{pifont}
\usepackage[utf8]{inputenc} 
\usepackage[T1]{fontenc}    
\usepackage{hyperref}       
\usepackage{url}            
\usepackage{booktabs}       
\usepackage{amsfonts}       
\usepackage{nicefrac}       
\usepackage{microtype}      
\usepackage{lipsum}
\usepackage{fancyhdr}       
\usepackage{graphicx}       
\graphicspath{{media/}}     

\title{DGEKT: A Dual Graph Ensemble Learning Method for Knowledge Tracing
}

\author{
  ChaoranCui, YumoYao, ChunyunZhang, HeboMa  \\
  Shandong University of Finance and Economics \\
  Jinan, China\\
  \texttt{crcui@sdufe.edu.cn} \\
   \And
  YulingMa \\
  Shandong Jianzhu University \\
  Jinan, China\\
  \texttt{mayuling@mail.sdu.edu.cn} \\
   \AND
   ZhaochunRen \\
   Shandong University \\
  QingDao, China\\
   \texttt{zhaochun.ren@sdu.edu.cn} \\
   \And
   Chen~Zhang \\
   Hong Kong Polytechnic University \\
  Hong Kong, China\\
   \texttt{jason-c.zhang@polyu.edu.hk} \\
   \And
   James~Ko \\
   Education University of Hong Kongn \\
  Hong Kong, China\\
   \texttt{jamesko@eduhk.hk} \\
}
\begin{document}
\maketitle

\begin{abstract}
Knowledge tracing aims to trace students' evolving knowledge states by predicting their future performance on concept-related exercises. Recently, some graph-based models have been developed to incorporate the relationships between exercises to improve knowledge tracing, but only a single type of relationship information is generally explored. In this paper, we present a novel Dual Graph Ensemble learning method for Knowledge Tracing (DGEKT), which establishes a dual graph structure of students' learning interactions to capture the heterogeneous exercise-concept associations and interaction transitions by hypergraph modeling and directed graph modeling, respectively. To ensemble the dual graph models, we introduce the technique of online knowledge distillation, due to the fact that although the knowledge tracing model is expected to predict students’ responses to the exercises related to different concepts, it is optimized merely with respect to the prediction accuracy on a single exercise at each step. With online knowledge distillation, the dual graph models are adaptively combined to form a stronger teacher model, which in turn provides its predictions on all exercises as extra supervision for better modeling ability. In the experiments, we compare DGEKT against eight knowledge tracing baselines on three benchmark datasets, and the results demonstrate that DGEKT achieves state-of-the-art performance.
\end{abstract}


\section{Introduction}
According to the statistics of UNESCO\footnote{https://en.unesco.org/covid19/educationresponse/globalcoalition}, more than 1.5 billion students worldwide have been affected by school and university closures due to the COVID-19 pandemic.
In response to this crisis, online learning, which serves as an alternative to traditional face-to-face learning, is developing on an unprecedented scale.
It is increasingly accessible for students to study on all sorts of online learning platforms, including Coursera, Udacity, and edX.

In the meantime, it has become crucial for online learning platforms to automatically estimate students' \emph{knowledge states}, i.e., their mastery level of different knowledge concepts (e.g., the concepts of linear equation, power function, and inequality in mathematics)~\cite{huang2020learning}.
This enables a personalized learning experience to be provided for each student, such as learning resource recommendation~\cite{wang2022personalized}, adaptive testing~\cite{zhao2018automatically}, and student profiling~\cite{zong2020behavior}.

\begin{figure}
    \centering
    \includegraphics[width=0.95\textwidth]{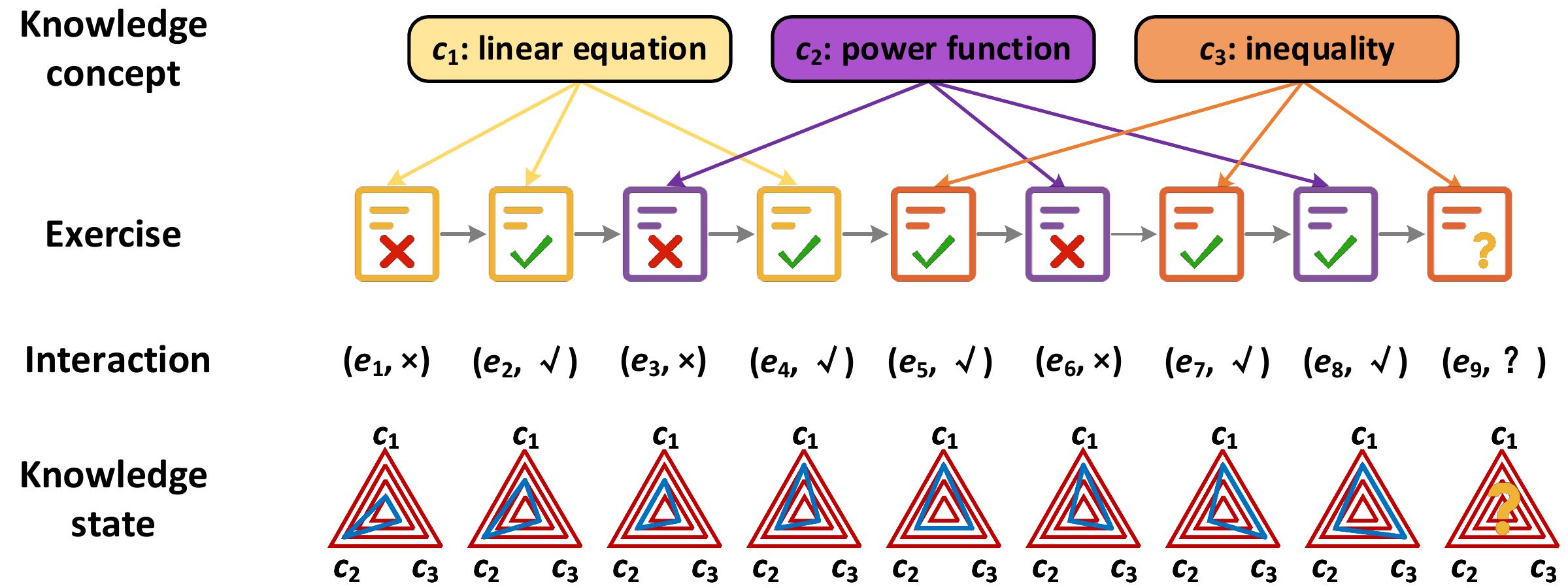}
    \caption{A simple schematic diagram of knowledge tracing.
    Based on a student's learning sequence of concept-related exercises, knowledge tracing predicts her/his responses to the following exercises and indirectly models the process by which the student's knowledge state of each concept changes, which is represented by a series of radar maps.}
    \label{fig:illustration}
\end{figure}

Knowledge tracing is one of the major research directions to address this problem.
Although the mission is to monitor students' evolving knowledge states during the learning process, it is difficult to dynamically obtain the true knowledge states of students~\cite{liu2021survey}.
As an alternative, knowledge tracing predicts students' future performance on concept-related exercises based on their past interactions.
In this way, students' knowledge states of each concept can be indirectly traced by observing their answers on the exercises related to the concept.
As illustrated in Fig.~\ref{fig:illustration}, knowledge tracing models take the exercise-answering sequence of a student as the input.
Generally, most models simplify the learning environments by assuming that the student's answers are only binary (i.e., correct or incorrect responses).
Given a new exercise, the output of knowledge tracing models is the probability that the student correctly answers the exercise.

In the literature, knowledge tracing has been investigated for decades.
At the early stage, some studies adopted the hidden Markov model to capture how students' knowledge state changes, among which the most popular method is bayesian knowledge tracing (BKT)~\cite{corbett1994knowledge,khajah2016deep}.
With the rise of deep learning, deep knowledge tracing (DKT) introduced recurrent neural networks (RNNs) to knowledge tracing and achieved promising results due to its ability to model long dependencies between learning interactions~\cite{piech2015deep}.
After that, more types of neural networks were used to improve knowledge tracing, such as memory-augmented neural networks~\cite{zhang2017dynamic,abdelrahman2019knowledge} and transformer networks~\cite{pandey2019self,ghosh2020context}.
Recently, motivated by the observation that knowledge concepts are not isolated but interrelated to each other, the relationship information was incorporated as an inductive bias to improve knowledge tracing.
Especially with the popularity of graph convolutional networks (GCNs)~\cite{Kipf2017semi}, different exercises  and their dependencies are typically viewed as nodes and edges in a graph, and the influence between exercises is modeled via the node representation learning applied on the graph.


In spite of the encouraging progress, graph-based knowledge tracing methods remain unsatisfactory, and the technology is still in its infancy.
A major limitation of existing research is the focus on a single type of relationship information, and the oversight of the fact that exercises are broadly connected via a variety of relationships.
The potential benefits of integrating different relationship information for knowledge tracing are under-explored.
In addition, only the exercise embeddings are generated through GCNs in existing research, which need to be further concatenated with the corresponding answer embeddings to represent students' learning interactions before they are fed into knowledge tracing models~\cite{nakagawa2019graph,liu2020improving}.
As pointed out in~\cite{piech2015deep}, the separate embeddings of exercises and answers may be incompatible and degrade the performance.

\begin{figure}
    \centering
    \includegraphics[width=0.6\textwidth]{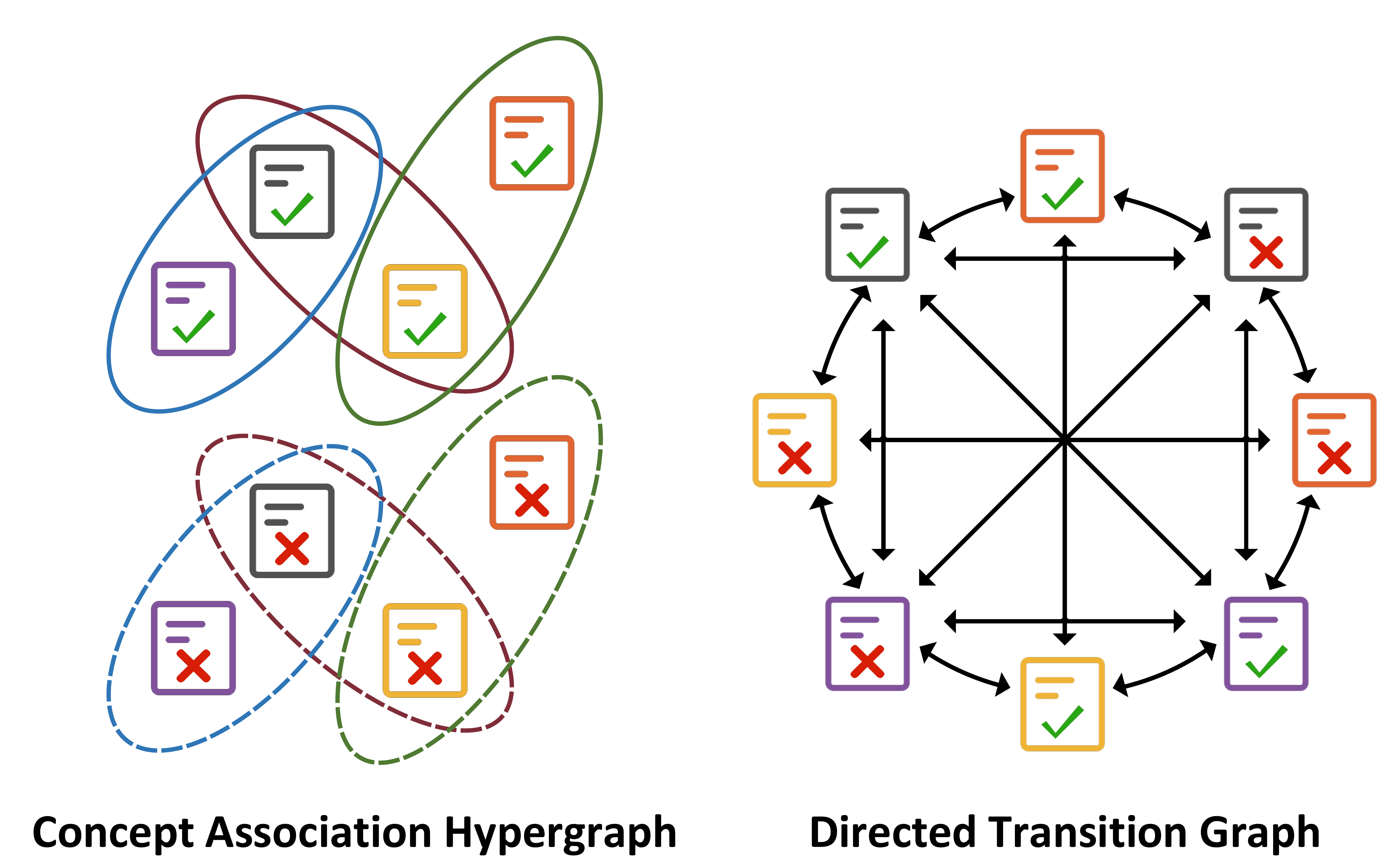}
    \caption{The dual graph structure of learning interactions, which consists of a concept association hypergraph (CAHG) revealing the associated knowledge concepts of learning interactions, as well as a directed transition graph (DTG) describing the directed transitions from one learning interaction to another in students' learning sequences.}
    \label{fig:dual_graph}
\end{figure}

In this paper, distinguished from prior studies relying on the dependencies between exercises, we investigate the relational graph of students' learning interactions, in which each node directly represents the combination of which exercise is answered and whether it is answered correctly or not.
As a result, the intermediate step of separately acquiring and integrating the exercise and answer embeddings is skipped.
More importantly, we establish a dual graph structure of learning interactions to capture their heterogeneous relationships.
As shown in Fig.~\ref{fig:dual_graph}, a concept association hypergraph (CAHG) is constructed to reveal the associated knowledge concepts of learning interactions.
Intuitively, a group of learning interactions can be made due to the mastery or non-mastery of the same knowledge concept.
In the hypergraph~\cite{chen2020neural}, a hyperedge expresses a group-wise relationship linking multiple nodes simultaneously.
Meanwhile, a directed transition graph (DTG) is used to describe the directed transitions from one learning interaction to another according to the learning sequences of all students.
This is inspired by the observation that owing to the relatively short time interval between them, two consecutive interactions are more likely to be triggered in students' knowledge states with little change.
In this way, the underlying collaborative effects across students are implicitly exploited.
By propagating and aggregating the information over the dual graph structure, the hypergraph convolutional networks~\cite{feng2019hypergraph} and directed graph convolutional networks~\cite{shi2019skeleton} are deployed to learn the node embeddings, respectively.
Furthermore, the hidden representations of students' knowledge states can be expressed in CAHG and DTG from two independent perspectives.

For a comprehensive framework, it is logical to combine the dual graph models.
The most straightforward way may be the concatenation- or adding-based fusion of their outputs.
But in this study, we innovatively introduce the idea of online knowledge distillation~\cite{zhang2018deep} to achieve this end.
Recall that knowledge tracing models aim to assess a student's knowledge state by predicting the student's performance on the exercises related to different concepts.
However, due to the use of the student's unique historical learning sequence, knowledge tracing models can only be trained with respect to the prediction accuracy of the response to a single exercise at each step.
With online knowledge distillation, we regard the dual graph models as peer student models and adaptively ensemble them to build a stronger teacher model using a gating mechanism~\cite{lan2018knowledge}.
Both student and teacher models are trained as usual by optimizing the prediction on the next exercise.
Additionally, the teacher's predictions on \emph{all exercises} are distilled back to student models as extra supervision, leading to more powerful modeling capacity.
During training, the dual graph models can exchange their information and learn from each other via the ensemble teacher model;
in testing, the stronger teacher model is deployed to achieve better prediction accuracy.


In a nutshell, we present a Dual Graph Ensemble learning method for Knowledge Tracing (DGEKT), which makes three main contributions:
\begin{itemize}
  \item We propose a novel dual graph structure of students' learning interactions for knowledge tracing, based on which DGEKT captures the heterogeneous exercise-concept associations and interaction transitions via hypergraph modeling and directed graph modeling, respectively.
  \item We leverage online knowledge distillation to form an ensemble teacher model by adaptively combining the dual graph models, which in turn provides its predictions on all exercises as extra supervision for higher modeling capacity.
        To the best of our knowledge, DGEKT is the first effort to incorporate knowledge distillation techniques to knowledge tracing.
  \item We conduct extensive experiments on three benchmark datasets in comparison with eight previous methods.
      The experimental results demonstrate that DGEKT achieves state-of-the-art performance for knowledge tracing.
\end{itemize}

The remainder of the paper is structured as follows.
Section~\ref{sec:related_work} reviews the related work.
Section~\ref{sec:method} presents the proposed method for knowledge tracing.
Experimental results and analysis are reported in Section~\ref{sec:experiment}, followed by the conclusion in Section~\ref{section:conclusion}.

\section{Related Work}\label{sec:related_work}
In this section, we first extensively review the existing literature on knowledge tracing.
Then, we present a brief overview of two fields of graph convolutional networks and knowledge distillation, which are closely related to our work.

\subsection{Knowledge Tracing}
Due to its great potential to support intelligent educational services, knowledge tracing has emerged as a hot research topic in recent decades.
In general, prior studies on knowledge tracing can be grouped into three categories: traditional non-deep learning methods, deep sequential modeling methods, and deep graph-based methods.

\subsubsection{Traditional Methods}
As the pioneering work for knowledge tracing, BKT~\cite{corbett1994knowledge} was proposed based on the hidden Markov model that used binary variables to represent students' knowledge states, i.e., whether a knowledge concept is mastered or not.
BKT considered four factors affecting students' responses: the initial knowledge level, learning rate, guessing probability, and slipping probability.
After that, many researchers extended BKT by introducing other important factors, such as students' individual variation in ability~\cite{khajah2016deep}, the difficulty level of exercises~\cite{Shen2022Assessing}, and prerequisite hierarchies and relationships within knowledge concepts~\cite{kaser2017dynamic}.
Another typical kind of traditional methods is logistic models~\cite{Pavlik2009Performance}, which directly adopted logistic functions to predict the probability of correctly answering exercises according to the side information about students and knowledge concepts.
Furthermore, knowledge tracing machines (KTM)~\cite{vie2019knowledge} was developed to take advantage of factorization machines~\cite{chen2020efficient} to generalize logistic models to characterize the pairwise feature interactions.
Despite the initial breakthrough of traditional methods, they mostly relied on linear models with restricted functional forms~\cite{piech2015deep}, which inevitably hindered the advancement of knowledge tracing.

\subsubsection{Sequential Modeling Methods}
With the huge surge of deep learning, different types of neural networks have been explored to model the learning sequences of students for knowledge tracing.
DKT~\cite{piech2015deep} was the first study that used the hidden state of RNNs at each step to represent students' evolving knowledge state and achieved much improved results over traditional methods.
Unlike DKT modeling knowledge states via a single vector, dynamic key-value memory networks (DKVMN)~\cite{zhang2017dynamic} introduced external memory matrices to store the knowledge concepts and update the corresponding knowledge mastery of students.
Due to the popularity of the attention mechanism~\cite{li2017neural} in deep learning, Pandey et al.~\cite{pandey2019self} proposed a self-attentive model for knowledge tracing (SAKT), which applied the self-attention mechanism to identify the past learning interactions relevant to the exercise to be answered, and made the prediction by focusing on these relevant ones.
Similarly, Ghosh et al.~\cite{ghosh2020context} presented a context-aware attentive knowledge tracing (AKT) model, which used the transformer network~\cite{vaswani2017attention} to capture the long-term dependencies between students' learning interactions, and designed a monotonic attention mechanism with an exponential decay curve to reduce the importance of interactions in the distant past.
Besides, some studies integrated deep sequence models with more elaborate factors related to learning, including the text content of exercises~\cite{liu2019ekt}, temporal cross-effects between knowledge concepts~\cite{wang2021temporal}, and students' individual cognition and knowledge acquisition sensitivity to exercises~\cite{long2021tracing}.

\subsubsection{Graph-based Methods}
It is well recognized that different exercises are not isolated but correlated to each other~\cite{liu2021survey}.
Therefore, some graph-based learning methods were developed to incorporate the relationship information to improve knowledge tracing.
For example, Nakagawa et al.~\cite{nakagawa2019graph} presented the graph-based knowledge tracing (GKT), which established the graph between knowledge concepts in a statistics- or learning-based manner and modeled students' proficiency in knowledge concepts with GCNs by formulating the task as a time-series node-level classification problem.
Deep hierarchical knowledge tracing (DHKT)~\cite{wang2019deep} augmented DKT by separately learning exercise and concept embeddings with an auxiliary task of predicting the associations between exercises and knowledge concepts.
As a further step, Liu et al.~\cite{liu2020improving} proposed a pre-training model named PEBG that learned the pre-training embeddings by constructing the exercise-concept bipartite graph and simultaneously exploiting the exercise-concept relationships, inter-exercise similarities, and inter-concept similarities.
In addition, Tong et al.~\cite{tong2020structure} proposed a structure-based knowledge tracing (SKT) framework, which exploited the prerequisite and similarity relations in the knowledge structure of concepts and considered both the temporal effect on the learning sequence and the spatial effect on the knowledge structure.
Despite the impressive progress, these graph-based methods performed the independent learning of exercise and answer embeddings, which may be incompatible when combined into the representations of learning interactions.
Moreover, early works encoded the graph structure between exercises or knowledge concepts depending on a single type of relationship information.
By contrast, we establish the dual graph structure of students' learning interactions, based on which the heterogeneous exercise-concept associations and interaction transitions are jointly captured.

\subsection{Graph Convolutional Networks}
Graph convolutional networks (GCNs) have emerged recently as a powerful tool for analyzing non-Euclidean data structure such as social networks in the era of deep learning~\cite{zhao2021Bilateral}.
The basic idea of GCNs is to transform, propagate, and aggregate the information from neighboring nodes for node representation learning.
Most GCNs fall into one of two families, i.e., spectral networks~\cite{Kipf2017semi} and spatial networks~\cite{hamilton2017inductive}.
The former defines convolution on graphs via the eigen decomposition of the graph Laplacian, while the latter defines graph convolution as a localized averaging operation with iteratively learned weights.
In traditional GCNs, only the pairwise connections between data were considered.
Hypergraph convolutional networks (HGCNs)~\cite{feng2019hypergraph} were further developed to generalize GCNS over hypergraphs that describe the topological structure involving non-pairwise relationships among nodes.
On the other hand, directed graph convolutional networks (DGCNs)~\cite{shi2019skeleton,zhang2021magnet} extend GCNs to directed graphs by discriminating between incoming and outgoing neighbors of each node during information propagation.
In our study, building upon the observation that exercises belonging to the same knowledge concept can naturally be viewed as a collective group, we leverage HGCNs in modeling the exercise-concept associations and avoid the information loss caused by decomposing the group-wise relationships into pairwise ones.
Besides, we adopt DGCNs to capture the transitions between learning interactions, which have not been fully investigated in prior research, and especially take account of the transition directions.

\begin{figure}
    \centering
    \includegraphics[width=1.0\textwidth]{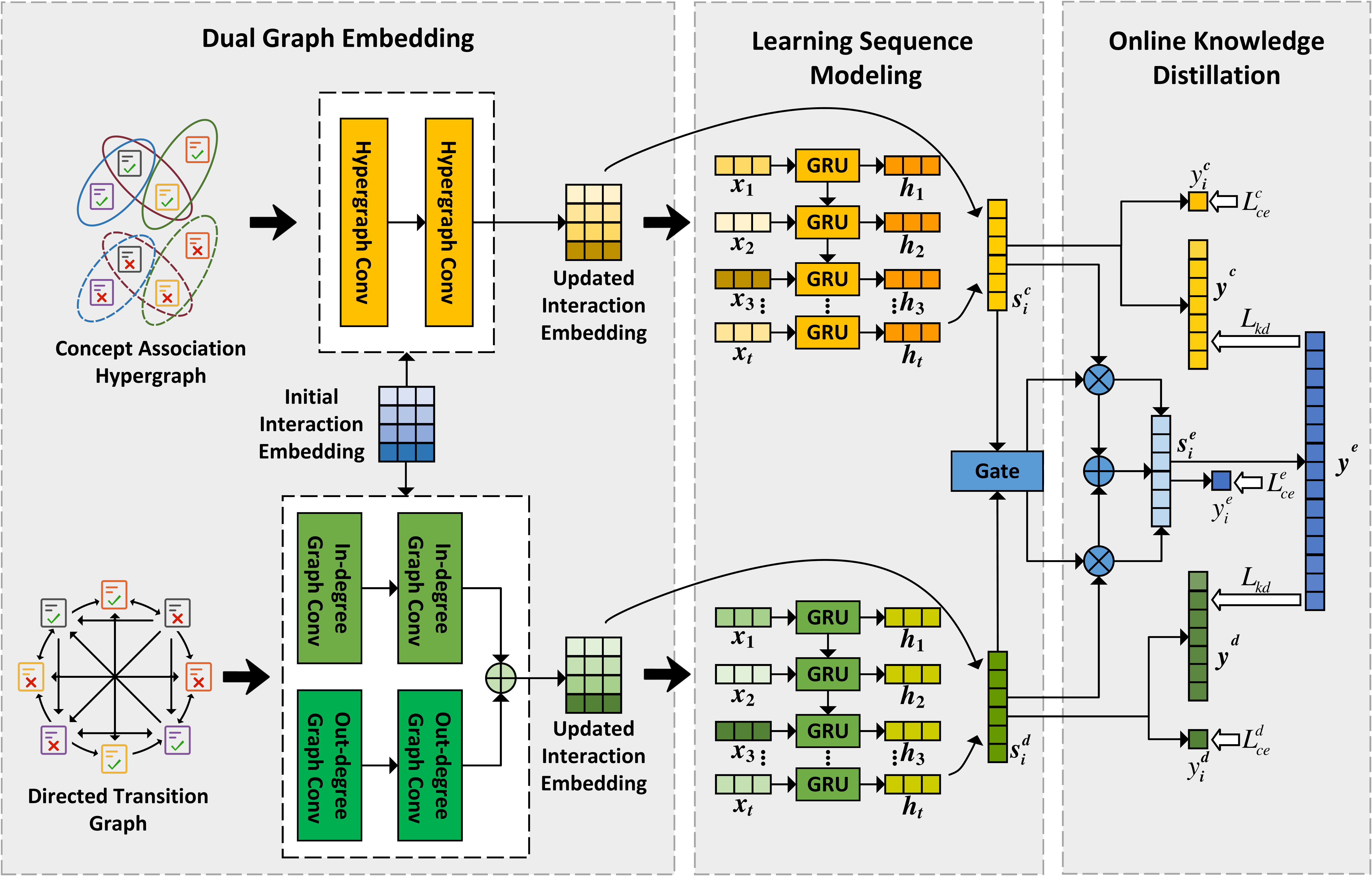}
    \caption{The overall architecture of the proposed DGEKT method.}
    \label{fig:architecture}
\end{figure}

\subsection{Knowledge Distillation}
Knowledge distillation is an effective deep learning technique that has been widely used to transfer information from one network to another whilst training constructively~\cite{wang2021knowledge}.
Generally, the information is transferred in the form of response-based knowledge, feature-based knowledge, or relation-based knowledge~\cite{gou2021knowledge}.
Most knowledge distillation methods often take offline learning strategies~\cite{hinton2015distilling}, in which a teacher model needs to be first pre-trained before distillation, and it is then used to guide the training of a student model during distillation.
To overcome the limitation of requiring the pre-trained teacher model, online knowledge distillation~\cite{lan2018knowledge,zhang2018deep} has been proposed to train a group of student models simultaneously by making them learn from each other in a peer-teaching approach.
As opposed to straightforward concatenation- or adding-based fusion, we apply online knowledge distillation to ensemble the dual graph models.
Besides the reference to students' responses to a single exercise at the next step, the ensemble teacher model can provide its predictions on all exercises back to student models as extra supervision, leading to more powerful modeling ability.

\begin{table}[tbp]
\caption{Summary of key notations and definitions.}
\centering
\begin{tabularx}{0.75\textwidth}{l  X}
    \toprule
    \textbf{Notation} & \textbf{Definition} \\
    \midrule
    $\mathcal{E}, \mathcal{C}$ & Set of exercises and concepts \\ [2pt]
    $n, m$ & Number of exercises and concepts \\ [2pt]
    $e_i \in \mathcal{E}$, $c_j \in \mathcal{C}$ & Exercise and concept exemplars \\ [2pt]
    $\mathcal{I}$ & Exercise-answering sequence of a student \\ [2pt]
    $t$ & Number of steps in the sequence $\mathcal{I}$\\ [2pt]
    $\left( {{e_k},{r_k}} \right) \in \mathcal{I}$ & Response $r_k$ to exercise $e_k$ at step $k$ \\ [2pt]
    $\mathcal{V}$ & Node set consisting of all learning interactions and $\left|\mathcal{V}\right| = 2n$\\ [2pt]
    $v_i \in \mathcal{V}, \bm{x}_i$ & Node exemplar $v_i$ and its embedding $\bm{x}_i$ \\ [2pt]
    $\mathcal{G}_c = (\mathcal{V},\mathcal{E}_c)$ & Concept association hypergraph (CAHG) \\ [2pt]
    $\mathcal{H}_j \in \mathcal{E}_c$ & Concept hyperedge containing multiple nodes \\ [2pt]
    $\mathcal{G}_d = (\mathcal{V},\mathcal{E}_d)$ & Directed transition graph (DTG) \\ [2pt]
    $A^{(\rm{in})}_{i,j}, A^{(\rm{out})}_{i,j}$ & Transition probability from $v_j$ to $v_i$ and vice versa in DTG \\ [2pt]
    $\bm{s}^c_i, \bm{s}^d_i$ & Knowledge states regarding $e_i$ of the student models from CAHG and DTG \\ [2pt]
    $\bm{s}^e_i$ & Knowledge state regarding $e_i$ of the ensemble teacher model \\ [2pt]
    ${y}^c_i, {y}^d_i$ & Predicted probability of the student models for correctly answering $e_i$ at the next step \\ [2pt]
    ${y}^e_i$ & Predicted probability of the ensemble teacher model for correctly answering $e_i$ at the next step \\ [2pt]
    \bottomrule
\end{tabularx}
\label{tab:notation}
\end{table}

\section{Method}\label{sec:method}
In this section, we present a dual graph ensemble learning method, called DGEKT, for knowledge tracing.
Firstly, we formulate the problem of knowledge tracing.
Next, we detail the dual graph structure of students' learning interactions based on which the heterogeneous relationships between interactions are captured.
Then, we illustrate the process of dynamic modeling of students' learning sequences.
Finally, we introduce an online knowledge distillation strategy to combine the dual graph models.
Fig.~\ref{fig:architecture} displays the overall architecture of our method.
For clarity, Table~\ref{tab:notation} summarizes some key notations and definitions used throughout this paper.

\subsection{Problem Formulation}
In an online learning platform, there exists a set of exercises $\mathcal{E}$.
All students are asked to answer different exercises in order to acquire related knowledge.
Suppose the exercise-answering sequence of a student is represented as ${\mathcal{I}} = \left\{ {\left( {{e_1},{r_1}} \right),\left( {{e_2},{r_2}} \right), \ldots ,\left( {{e_t},{r_{t}}} \right)} \right\}$, where $e_k \in \mathcal{E}$ denotes the exercise the student answers at step $k$, $t$ is the total number of steps, and
$r_k$ denotes the student's response to $e_k$.
Similar to most previous studies on knowledge tracing, we simplify $r_k$ as a binary value, that is, $r_k=1$ if the student correctly answers $e_k$; otherwise, $r_k$ equals zero.

Given the sequences of students' learning interactions, knowledge tracing seek to monitor students' evolving knowledge states by predicting their performance on the following exercises.
Therefore, it is formulated to predict the probability that a student will give the correct response to the next exercise, i.e., $p\left( {{r_{t + 1}} = 1\left| {{e_{t + 1}},{\mathcal{I}}} \right.} \right)$.

\subsection{Dual Graph Embedding}
In this paper, we incorporate the potential relationship information to improve knowledge tracing, but differ from prior research~\cite{nakagawa2019graph,liu2020improving} in establishing the relational graph of learning interactions instead of exercises.
Specifically, assume there are totally $n$ exercises.
$\mathcal{V} = \left\{ {v_{1+} ,v_{1-} ,v_{2+} ,v_{2-} , \ldots ,v_{n+} ,v_{n-} } \right\}$ denotes the graph nodes, in which the nodes $v_{i+}$ and $v_{i-}$ represent the interactions that students correctly and incorrectly answer the exercise $e_i \in \mathcal{E}$, respectively.
Obviously, the number of nodes is twice the number of exercises.
$v_{i+}$ and $v_{i-}$ will be projected into two embeddings $\bm{x}_{i+}$ and $\bm{x}_{i-}$, which can be understood as the knowledge levels underlying students' right and wrong responses to $e_i$, respectively.
In this way, we eliminate the necessity of having separate representations of exercises and answers in previous studies~\cite{piech2015deep}.

It is intuitively plausible that closely related learning interactions are more likely to reflect students' similar knowledge levels.
Therefore, we establish a dual graph structure of learning interactions to capture the heterogeneous relationships between them, consisting of a concept association hypergraph (CAHG) and a directed transition graph (DTG).
Then, hypergraph convolutional networks~\cite{feng2019hypergraph} and directed graph convolutional networks~\cite{shi2019skeleton} are deployed to learn the node embeddings over CAHG and DTG, respectively.
The detailed procedures are described as follows:

\subsubsection{Concept Association Hypergraph}
In knowledge tracing, each exercise is generally related to at least one knowledge concept $c_j \in \mathcal{C}$, where $\mathcal{C} = \left\{ {c_{1} ,c_{2} , \ldots , c_{m} } \right\}$ denotes the set of knowledge concepts with size $m$.
The exercise-concept associations have been typically encoded as pairwise connections between exercises and concepts~\cite{liu2020improving} or between exercises alone~\cite{nakagawa2019graph}.
However, we argue that different exercises can naturally be viewed as a collective group if they are associated with the same concept, and simply decomposing the group-wise relationships into pairwise ones may cause the loss of information.
From this point, we introduce the hypergraph structure CAHG to characterize the group-wise exercise-concept associations.
Hypergraph is an extension of simple graph by defining a hyperedge as a subset of nodes.
Let ${\mathcal{G}_c} = \left( {\mathcal{V},{\mathcal{E}_c}} \right)$ denote CAHG, where ${\mathcal{E}_c} = \left\{ {{\mathcal{H}_{1 + }},{\mathcal{H}_{1 - }},{\mathcal{H}_{2 + }},{\mathcal{H}_{2 - }}, \ldots ,{\mathcal{H}_{m + }},{\mathcal{H}_{m - }}} \right\}$ is the set of hyperedges.
Recall that each exercise corresponds to two interaction nodes of correctly and incorrectly answering the exercise.
Analogously, each knowledge concept $c_j$ is decoupled into two hyperedges $\mathcal{H}_{j+}$ and $\mathcal{H}_{j-}$, making the cardinality of ${\mathcal{E}_c}$ to be twice that of $\mathcal{C}$.
The exercise-concept associations can be indicated by the incidence relationships between the hyperedges and their nodes.
For example, if an excise $e_i$ is associated with the concept $c_j$, students' right and wrong responses to $e_i$ are probably made due to their mastery and non-mastery of $c_j$, leading to $v_{i+} \in \mathcal{H}_{j+}$ and $v_{i-} \in \mathcal{H}_{j-}$ at the same time.
Due to the existence of correspondences between exercises and concepts, there are many-to-many correspondences between nodes and hyperedges.
In the following, we shall omit the subscript $+$ or $-$ when no confusion can arise.

In the hypergraph, $\mathcal{U}_i = \left\{ {\mathcal{H}_j\left| {{v_i} \in \mathcal{H}_j} \right.} \right\}$ denotes the subset of hyperedges containing node $v_i$, based on which the degree of $v_i$ can be defined as $d_i = \left| \mathcal{U}_i \right|$.
Besides, the degree of hyperedge $\mathcal{H}_j$ is equal to the number of nodes within it, i.e., $g_j = \left| \mathcal{H}_j \right|$.
To learn the embedding $\bm{x}_i$ of $v_i$, hypergraph convolutional networks~\cite{feng2019hypergraph} can be applied over CAHG, which defines an information propagation rule in hypergraphs via a convolution operator in each layer.
Formally, the convolution operator updates $\bm{x}_i$ by aggregating the information from $v_i$ itself as well as the local neighbors in each hyperedge to which $v_i$ belongs:
\begin{equation}\label{eq:hgcn}
\bm{x}_i^{(l)} = \phi \left( \sum\limits_{{\mathcal{H}_j} \in \mathcal{U}_i} \frac{1}{g_j}{\sum\limits_{{v_q} \in {\mathcal{H}_j}} {\frac{1}{\sqrt{d_id_q}}{\bm{\Theta}^{(l)}}\bm{x}_q^{(l-1)}} }\right),
\end{equation}
where $\bm{x}_i^{(l-1)}$ and $\bm{x}_i^{(l)}$ are the input and output embeddings of $v_i$ in the $l$-th layer, $\bm{\Theta}^{(l)}$ is a learnable parameter matrix, and $\phi(\cdot)$ denotes a nonlinear activation function like LeakyReLU~\cite{maas2013rectifier}.
Intuitively, the updating smoothes the embeddings of nodes locally linked by the same hyperedges.

\subsubsection{Directed Transition Graph}
Except for exercise-concept associations, the rich transitions between interactions in students' learning sequences are also an important factor in knowledge tracing, which, however, has not been fully explored in prior research~\cite{nakagawa2019graph}.
Intuitively, owing to the relatively short time interval between them, two consecutive interactions of students are more likely to be triggered in the knowledge states with little change.
In light of this, we adopt the directed graph structure DTG for modeling the transitions between students' learning interactions.
We denote DTG by ${\mathcal{G}_d} = \left( {\mathcal{V},{\mathcal{E}_d}} \right)$, where the set of edges ${\mathcal{E}_d} \subseteq \mathcal{V} \times \mathcal{V}$ contains the edge $\left( {{v_i},{v_j}} \right)$ if the interaction $v_j$ comes right after the interaction $v_i$ in any student's learning sequence.
Furthermore, two transition probability matrices $\bm{A}^{(\rm{in})}$ and $\bm{A}^{(\rm{out})}$ are defined to assign weights to edges, i.e.,
\begin{equation}\label{eq:in_transition}
A^{(\rm{in})}_{i,j} = \left\{ \begin{aligned}
&\frac{{{n_{j,i}}}}{{\sum\nolimits_k {{n_{k,i}}} }}&&\mathrm{if}\ \left( {{v_j},{v_i}} \right) \in \mathcal{E}_d;\\
&0 &&\mathrm{otherwise},
\end{aligned} \right.
\end{equation}
and
\begin{equation}\label{eq:out_transition}
A^{(\rm{out})}_{i,j} = \left\{ \begin{aligned}
&\frac{{{n_{i,j}}}}{{\sum\nolimits_k {{n_{i,k}}} }}&&\mathrm{if}\ \left( {{v_i},{v_j}} \right) \in \mathcal{E}_d;\\
&0 &&\mathrm{otherwise},
\end{aligned} \right.
\end{equation}
where $n_{i,j}$ denotes the number of times $v_j$ occurs immediately after $v_i$ in the learning sequences of all students.
In this way, the historical learning records of different students can be implicitly leveraged in collaboration.
We also include self-loops in DTG and set $A^{(\rm{in})}_{i,i} =1$ and $A^{(\rm{out})}_{i,i} =1$ for each node $v_i \in \mathcal{V}$.
The in-degree and out-degree of $v_i$ are computed as $d^{(\rm{in})}_{i} = {\sum\nolimits_k {A^{(\rm{in})}_{i,k}} }$ and $d^{(\rm{out})}_{i} = {\sum\nolimits_k {A^{(\rm{out})}_{i,k}} }$, respectively.

In DTG, we deploy directed graph convolutional networks~\cite{shi2019skeleton,zhang2021magnet} for node representation learning.
Particularly, we propose to update the embedding $\bm{x}_i$ of node $v_i$ via the bi-directional information propagation in parallel:
one diffuses the information along the direction of edges in the graph, so that the node embedding can be updated by receiving knowledge from the ancestor nodes:
\begin{equation}\label{eq:in_update}
\overline{\bm{x}}_i^{(l)} = \phi \left( {\sum\limits_{{v_j} \in \mathcal{N}_i^{\rm{(in)}} \cup \left\{ {{v_i}} \right\}} {\frac{{A_{i,j}^{\rm{(in)}}}}{{\sqrt {{d_i^{\rm{(in)}}}{d_j^{\rm{(out)}}}} }}{\bm{\Phi} ^{(l)}}\bm{x}_j^{(l-1)}} } \right);
\end{equation}
the other diffuses the information against the direction of edges in the graph, so that the node embedding can be updated by receiving knowledge from the descendant nodes:
\begin{equation}\label{eq:in_update}
\widetilde{\bm{x}}_i^{(l)} = \phi \left( {\sum\limits_{{v_j} \in \mathcal{N}_i^{\rm{(out)}} \cup \left\{ {{v_i}} \right\}} {\frac{{A_{i,j}^{\rm{(out)}}}}{{\sqrt {{d_i^{\rm{(out)}}}{d_j^{\rm{(in)}}}} }}{\bm{\Psi} ^{(l)}}\bm{x}_j^{(l-1)}} } \right).
\end{equation}
Here, $\mathcal{N}_i^{\rm{(in)}}$ and $\mathcal{N}_i^{\rm{(out)}}$ are the sets of ancestor and descendant nodes of $v_i$, respectively.
$\bm{\Phi}^{(l)}$ and $\bm{\Psi}^{(l)}$ are the model parameters to be learned in the $l$-th layer.
Finally, the output embedding ${\bm{x}}_i^{(l)}$ is generated by simply adding $\overline{\bm{x}}_i^{(l)} $ and $\widetilde{\bm{x}}_i^{(l)}$.
More complicated combination methods can be adopted, such as fusing $\overline{\bm{x}}_i^{(l)} $ and $\widetilde{\bm{x}}_i^{(l)}$ with a learnable multilayer perceptron, but this did not lead to significant performance improvement in our experiments.

\subsection{Learning Sequence Modeling}
After building the dual graph structure of learning interactions, the learning sequence of a student corresponds to a unique path in each graph, which can be further represented through sequential models.
In this study, we follow the well-known DKT network~\cite{piech2015deep} to accomplish this process, due to its relatively simpler structure and the ability to directly handle learning interactions.
Note that our method is also compatible with other recently proposed sequential modeling techniques for knowledge tracing~\cite{pandey2019self,ghosh2020context}.

Given the learning sequence of a student ${\mathcal{I}} = \left\{ {\left( {{e_1},{r_1}} \right),\left( {{e_2},{r_2}} \right), \ldots ,\left( {{e_t},{r_{t}}} \right)} \right\}$, each learning interaction $\left( {{e_k},{r_k}} \right) \in {\mathcal{I}}$ can be embedded to $\bm{x}_k$ as described above.
Similar to DKT using the vanilla RNN, we apply the gated recurrent unit (GRU) model to process the learning sequence.
At each step $k$, the GRU model executes the calculations as follows:
\begin{equation}\label{eq:GRU}
\begin{split}
\bm{r}_{k}&=\sigma\left(\emph{\textbf{W}}_{r}\left[\emph{\textbf{h}}_{k-1}, \emph{\textbf{x}}_{k}\right]+\textbf{b}_{r}\right),  \\  
\bm{u}_{k}&=\sigma\left(\emph{\textbf{W}}_{u}\left[\emph{\textbf{h}}_{k-1}, \emph{\textbf{x}}_{k}\right]+\textbf{b}_{u}\right),  \\
\bm{c}_{k}&=\tanh \left(\emph{\textbf{W}}_c\left[\bm{r}_{k} \odot \emph{\textbf{h}}_{k-1}, \emph{\textbf{x}}_{k}\right]+\textbf{b}_{c}\right), \\
\bm{h}_{k}&=\left(\bm{1}-\bm{u}_{k}\right) \odot \emph{\textbf{h}}_{k-1}+\bm{u}_{k} \odot \bm{c}_{k},
\end{split}
\end{equation}
where $\bm{h}_{k-1}$ is a hidden state summarizing the information of past steps.
$\bm{h}_{k-1}$ and $\bm{x}_k$ are concatenated and transformed into a reset gate $\bm{r}_k$ and an update gate $\bm{u}_k$, respectively.
The former determines how much the past information should be forgotten, whereas the latter controls how much the information needs to be brought into the current hidden state $\bm{h}_k$.
Then, $\bm{h}_{k-1}$ is reset with $\bm{r}_k$, and again concatenated with $\bm{x}_k$ to generate a memory cell $\bm{c}_k$.
$\bm{c}_k$ represents the new information to be added to $\bm{h}_k$.
Finally, $\bm{h}_k$ is computed as the combination of $\bm{h}_{k-1}$ and $\bm{c}_k$, and $\bm{u}_k$ serves as a balance factor in this procedure.
In Eq.~\eqref{eq:GRU}, $\textbf{W}_{r}$, $\textbf{W}_{u}$, and $\textbf{W}_{c}$ are the transformation matrices, and $\textbf{b}_{r}$, $\textbf{b}_{u}$, and $\textbf{b}_{c}$ are the bias terms to be learned.
$\odot$ denotes the Hadamard product, and $\sigma(\cdot)$ and $\tanh(\cdot)$ denote the sigmoid and tanh activation functions, respectively.

In DKT, the latest hidden state $\bm{h}_t$ is considered to reflect a student's knowledge state.
Given an exercise $e_i \in \mathcal{E}$,  we further concatenate $\bm{h}_t$ with the node embeddings $\bm{x}_{i+}$ and $\bm{x}_{i-}$ to represent the student's knowledge state regarding $e_i$:
\begin{equation}\label{eq:state_exercise}
\bm{s}_{i} = \left[\bm{h}_t, \bm{x}_{i+}, \bm{x}_{i-}\right],
\end{equation}
which comprehensively considers $\bm{h}_t$ and the knowledge levels for answering $e_i$ correctly and incorrectly.
$\bm{s}_{i}$ is then mapped to the probability that the student will give the correct response to $e_i$ at the next step $t+1$:
\begin{equation}\label{eq:prediction}
{y}_i = \sigma ({z_i}) = \frac{1}{{1 + {e^{ - {z_i}}}}},
\end{equation}
where
\begin{equation}\label{eq:logit}
{z_i} = {\bm{w}_o}^T{\bm{s}_{i}} + {b_o}
\end{equation}
is the logit or unnormalized log probability~\cite{szegedy2016rethinking}, and $\bm{w}_o$ and $b_o$ are the readout weight and bias parameters, respectively.
To jointly learn all model parameters, a cross-entropy loss between the predicted probability ${y}_i$ and the true response $r_{t+1}$ can be optimized:
\begin{equation}\label{eq:ce_loss}
{{\mathcal{L}}_{ce}} = - {\sum\limits_{i = 1}^n {\delta _{t + 1}^i({r_{t + 1}}\log y_i}}+(1 - {r_{t + 1}})\log (1 - y_i)),
\end{equation}
where $\delta _{t + 1}^i$ is a Dirac delta returning 1 when $e_i$ is the exercise the student actually did at the next step, and 0 otherwise.

\subsection{Online Knowledge Distillation}
Following the above procedures, we can separately model the learning sequence of a student with the node embeddings of CAHG and DTG, and obtain two independent representations of the knowledge states $\bm{s}^c_i$ and $\bm{s}^d_i$ as well as the predicted probabilities ${y}^c_i$ and ${y}^d_i$ for the exercise $e_i$.
To ensemble the two graph models, a natural thought is to directly combine $\bm{s}^c_i$ and $\bm{s}^d_i$ or ${y}^c_i$ and ${y}^d_i$~\cite{xia2020self,yu2021self}.
However, in this study, we introduce the idea of online knowledge distillation~\cite{zhang2018deep} for this purpose.
As shown in Eq.~\eqref{eq:ce_loss}, although the knowledge tracing model is expected to predict the students' responses to the exercises related to different concepts, it is optimized merely with respect to the prediction accuracy on a single exercise at the next step.
In light of this, we regard the two graph models as peer student models and ensemble them to build a stronger teacher model, which provides its predicted results on all exercises back to student models as extra supervision.

Specifically, due to the importance variety of student models, we adopt a gating mechanism~\cite{lan2018knowledge} to adaptively ensemble their generated knowledge states $\bm{s}^c_i$ and $\bm{s}^d_i$ to build the teacher model:
\begin{equation}\label{eq:gating_fusion}
\bm{s}_i^e = \bm{g} \odot \bm{s}_i^c + \left( {\bm{1} - \bm{g}} \right) \odot \bm{s}_i^d,
\end{equation}
where the gate $\bm{g}$ is given by
\begin{equation}\label{eq:gate}
\bm{g} =\sigma\left(\emph{\textbf{W}}_{g}\left[\bm{s}_i^c, \bm{s}_i^d\right]+\textbf{b}_{g}\right)
\end{equation}
with ${\textbf{W}}_{g}$ and $\textbf{b}_{g}$ being the learnable parameters.
Intuitively, $\bm{g}$ serves to balance the importance of the two student models.
Given $\bm{s}^c_i$, $\bm{s}^d_i$, and $\bm{s}^e_i$, their corresponding logits for the exercise $e_i$ can be yield from Eq.~\eqref{eq:logit} and denoted by $z^c_i$, $z^d_i$, and $z^e_i$, respectively.

To distill the knowledge of the ensemble teacher model back into student models, we firstly compute the soft version of their predicted probabilities at a temperature of $\gamma$:
\begin{equation}\label{eq:soft_probability}
\begin{split}
\tilde{y}_i^c &= \frac{1}{{1 + {e^{ - {z_i^c}/\gamma}}}},  \\
\tilde{y}_i^d &= \frac{1}{{1 + {e^{ - {z_i^d}/\gamma}}}},  \\
\tilde{y}_i^e &= \frac{1}{{1 + {e^{ - {z_i^e}/\gamma}}}}. \\
\end{split}
\end{equation}
Then, we design a distillation loss to encourage each student model matching with the teacher's predictions across all exercises:
\begin{equation}\label{eq:kd_loss}
{\mathcal{L}_{kd}} = \frac{1}{n}\sum\limits_{i = 1}^n {{{\left\| {\tilde{y}_i^e - \tilde{y}_i^c} \right\|}_1} + } {\left\| {\tilde{y}_i^e - \tilde{y}_i^d} \right\|_1}.
\end{equation}
Here, the discrepancy between the student and teacher models is measured by the $L_1$-norm distance of their outputs rather than the Kullback-Leibler (KL) divergence commonly used in knowledge distillation~\cite{hinton2015distilling}.
This is supported by the recent finding that the $L_1$-norm distance is a tighter bound on the classification error probability~\cite{wang2021re}.
Moreover, the distillation loss can also be viewed as a learned label smoothing regularization~\cite{yuan2020revisiting,muller2019does}, which makes the student and teacher models regularize the training process of each other and therefore improves model generalization.

Finally, both the student and teacher models need to be trained simultaneously with reference to the prediction performance of the student's response to the next exercise, according to the cross-entropy loss defined in Eq.~\eqref{eq:ce_loss}.
Therefore, we obtain the overall loss for training the proposed DGEKT as:
\begin{equation}\label{eq:overall_loss}
\mathcal{L} = \mathcal{L}_{ce}^c + \mathcal{L}_{ce}^d + \mathcal{L}_{ce}^e + \lambda {\mathcal{L}_{kd}},
\end{equation}
where $\mathcal{L}_{ce}^c$ and $\mathcal{L}_{ce}^d$ are the cross-entropy loss terms for the student models derived from CAHG and DTG, respectively, and $\mathcal{L}_{ce}^e$ is that for the ensemble teacher model.
$\lambda$ is a trade-off hyperparameter controlling the relative contributions of the cross-entropy and distillation losses.
Once the training is completed, the ensemble teacher model is deployed for achieving higher accuracy during testing.

\section{Experiment}\label{sec:experiment}
In this section, we describe the experimental setup details and report a series of experimental results to validate the effectiveness of the proposed DGEKT.

\begin{table}[tbp]
\caption{Dataset statistics.}
\centering
\begin{tabular}{l c c c}
    \toprule
    Dataset & ASSIST09 & ASSIST17 & EdNet \\
    \midrule
    \# Students & 4,151 & 1,709 & 5000 \\[2pt]
    \# Exercises & 16,891 & 3,162 & 10,795 \\[2pt]
    \# Concepts & 101 & 102 & 188 \\[2pt]
    \# Exercises per concept  & 167.24 & 37.98 & 148.73 \\[2pt]
    \# Concepts per exercise & 1.00 & 1.23 & 2.26 \\[2pt]
    \# Records & 274,590 & 942,816 & 225,625 \\[2pt]
    \% Correct answering records & 66.16\% & 37.26\% & 58.99\% \\[2pt]
    \% Incorrect answering records & 33.84\% & 62.74\% & 41.01\% \\[2pt]
    \bottomrule
\end{tabular}
\label{tab:dataset}
\end{table}

\subsection{Dataset}
We evaluate our method on three benchmark datasets for knowledge tracing, i.e., ASSIST09, ASSIST17, and EdNet.
The statistics of the three datasets are listed in Table~\ref{tab:dataset}.
On all datasets, the maximum length of students' learning sequences is set to 50.
If a student has more than 50 exercise-answering records, we partition his/her learning sequence into multiple subsequences.
We randomly select 80\% of students for training and validation and the remaining 20\% for testing.

\subsubsection*{ASSIST09\footnote{https://sites.google.com/site/assistmentsdata/home/2009-2010-assistment-data}}
The dataset is collected from the ASSISTments online tutoring platform\footnote{https://new.assistments.org} during the 2009-2010 school year.
We filter out the exercises without knowledge concepts and the students with less than three learning interactions.

\subsubsection*{ASSIST17\footnote{https://sites.google.com/view/assistmentsdatamining/data-mining-competition-2017}}
The dataset is obtained from the ASSISTments data mining competition 2017.
We do the same preprocessing as ASSIST09.

\subsubsection*{EdNet\footnote{https://github.com/riiid/ednet}}
The dataset is contributed by Choi et al. in~\cite{choi2020ednet}.
It is a large dataset with over 130 million records that involve about 0.78 million students.
For computational efficiency, we sample the learning records of 5,000 students from the dataset.

\subsection{Evaluation Metric}
The prediction of students' responses to exercises is considered as a binary classification task, i.e., predicting whether a student answers an exercise correctly or not.
As shown in Table~\ref{tab:dataset}, on account of the imbalance between the correct and incorrect answering records, we evaluate the classification performance of each algorithm in terms of the Area Under ROC Curve (AUC).
A higher value of AUC indicates better performance in predicting students' responses, and the value of 0.5 represents the performance of making predictions by random guessing.

\subsection{Competitor}
We compare the performance of DGEKT against eight knowledge tracing models in three categories, including:

\subsubsection{Traditional Methods}
These methods adopt non-deep learning techniques to predict students' responses.
\begin{itemize}
  \item \textbf{BKT}~\cite{corbett1994knowledge} is based on the hidden Markov model, and models students' knowledge states for each concept as a binary variable.
  \item \textbf{KTM}~\cite{vie2019knowledge} uses the factorization machine to encode the side information about exercises or students into the model.
For fairness, we only consider the exercise IDs and concept IDs as side information in comparison.
\end{itemize}

\subsubsection{Sequential Modeling Methods}
These methods address knowledge tracing with deep sequence models.
\begin{itemize}
  \item \textbf{DKT}~\cite{piech2015deep} applies RNN to process the learning sequences of students, and represents their knowledge states with the hidden states of RNN.
  \item \textbf{DKVMN}~\cite{zhang2017dynamic} introduces a memory-augmented neural network, which uses a key matrix to represent knowledge concepts and a value matrix to store and update the mastery of corresponding concepts of students over time.
  \item \textbf{SAKT}~\cite{pandey2019self} employs the self-attention mechanism to identify the past learning interactions relevant to the exercise to be answered, and makes the prediction by focusing on these relevant ones.
  \item \textbf{AKT}~\cite{ghosh2020context} uses the transformer network to model the learning sequences of students, and proposes a monotonic attention mechanism using an exponential decay curve to reduce the importance of interactions in the distant past.
\end{itemize}

\subsubsection{Graph-based Methods}
These methods incorporate the graph structure of exercises or knowledge concepts as a relational inductive bias to improve knowledge tracing.
\begin{itemize}
  \item \textbf{GKT}~\cite{nakagawa2019graph} learns the relationships between knowledge concepts and formulates knowledge tracing as a time-series node-level classification problem with GCNs.
  \item \textbf{PEBG}~\cite{liu2020improving} constructs the bipartite graph of exercise-concept relations to learn the pre-training embeddings of exercises, which are then used as the inputs to the DKT model.
      For fairness, the attribute features of exercises defined in~\cite{liu2020improving}, such as difficulty level, type, and average response time, were not exploited in our implementation.
  \item \textbf{DGEKT} is our proposed method in this paper.
\end{itemize}

\begin{table}[tbp]
\caption{Performance comparison on three benchmark datasets.}
\centering
\begin{threeparttable}
\setlength{\tabcolsep}{1.5mm}{
\begin{tabular}{c c c c c}
    \toprule
    Category & Method & ASSIST09& ASSIST17 &EdNet \\
    \midrule
    \multirow{2}{*}{Traditional} & BKT~\cite{corbett1994knowledge} & 72.05\%  & 63.65\% & 66.68\% \\ [2pt]
    & KTM~\cite{vie2019knowledge} & 72.92\%  & 67.32\% & \underline{69.79\%} \\ [2pt]
    \midrule
   \multirow{4}{*}{Sequential Modeling} & DKT~\cite{piech2015deep} & 73.37\%  & \underline{70.70\%} & 67.08\% \\ [2pt]
   & DKVMN~\cite{zhang2017dynamic} & 73.98\% & 66.70\% & 66.75\% \\ [2pt]
    & SAKT~\cite{pandey2019self} &  74.98\%  &  66.48\% & 63.45\% \\ [2pt]
    & AKT~\cite{ghosh2020context} & \underline{75.61\%}  & 70.34\% & 69.55\% \\ [2pt]
    \midrule
    \multirow{3}{*}{Graph-based} & GKT~\cite{nakagawa2019graph} & 73.96\%  & 69.16\% & 67.89\% \\ [2pt]
    & PEBG~\cite{liu2020improving} & 73.47\%  & 68.03\% & 66.94\% \\ [2pt]
    & DGEKT & \textbf{76.56\%}  &  \textbf{77.54\%} & \textbf{70.07\%} \\ [2pt]
    \bottomrule
\end{tabular}
}
\begin{tablenotes}
\footnotesize
\item The best result in terms of each metric is indicated in bold, and the second best one is underlined. This convention is also adopted in the following tables.
\end{tablenotes}
\end{threeparttable}
\label{tab:overall_performance}
\end{table}

\subsection{Implementation Details}
Our implementation determined the optimal hyperparameters of baseline methods either following the suggested settings in the original papers or based on the AUC value obtained on the validation set.
In the proposed DGEKT, all learning interactions were consistently embedded in a 256-dimensional space.
We adopted a two-layered graph network for node embedding on CAHG and DTG, respectively.
The number of hidden layers and the number of cells per layer in the GRU model were 1 and 128, respectively.
For online knowledge distillation, we set the temperature factor $\gamma = 0.5$ and the trade-off hyperparameter $\lambda = 0.01$, respectively.
We will discuss the sensitivity of DGEKT with regard to the changing of different hyperparameters later.

We implemented the network training and testing using the deep learning library Pytorch\footnote{https://pytorch.org}.
The network was trained with the mini-batch Adam optimizer~\cite{kingma2014adam}.
We set the batch size to 128.
The initial learning rate was $10^{-3}$ for all layers, and the maximum number of epochs was 600 during training.
For the sake of reproducibility, the codes and models of our work have been released at \url{https://github.com/Yumo216/DGEKT}.

\begin{figure}
    \centering
    \includegraphics[width=0.6\textwidth]{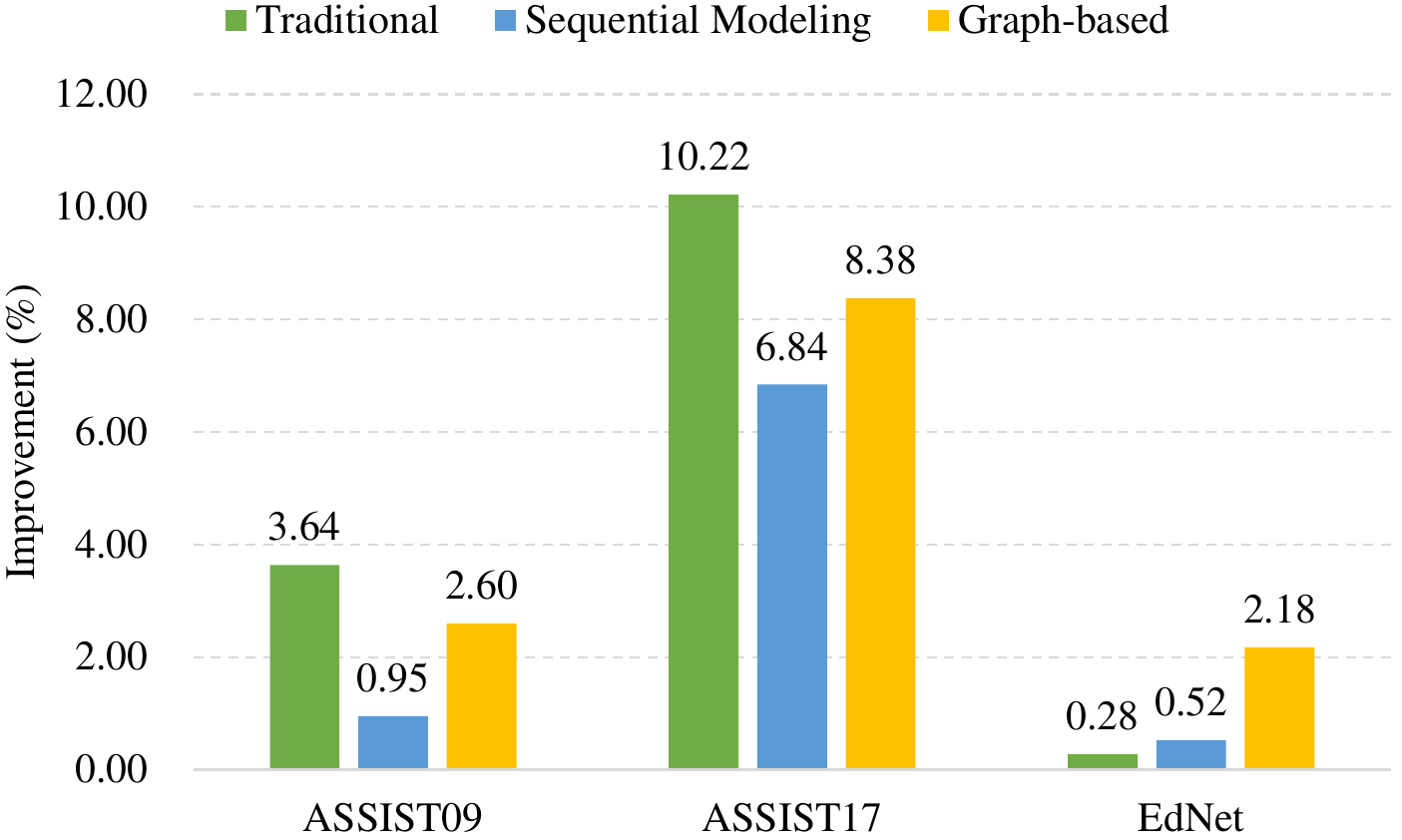}
    \caption{The improvement over baselines of different categories.}
    \label{fig:AUC}
\end{figure}

\subsection{Performance Comparison}
Table~\ref{tab:overall_performance} lists the performance of the proposed DGEKT compared to baseline methods.
We can observe that DGEKT outperforms all of the eight baselines and achieves the best performance on all datasets.
More precisely, it is better than the best baseline by 0.95\%, 6.84\%, and 0.28\% in terms of AUC on ASSIST09, ASSIST17, and EdNet, respectively.
It deserves to mention that DGEKT is equipped with a sequential modeling component similar to DKT, but it is significantly ahead of DKT.
In addition, even without the attention mechanism~\cite{vaswani2017attention} in sequential modeling, the performance of DGEKT is still consistently higher than the attention-based baselines, such as SAKT and AKT.
That means that introducing the modules of dual graph embedding and online knowledge distillation into DGEKT well benefits the performance of knowledge tracing.
This point will be further verified later.

Overall, traditional methods frequently fall behind recently proposed deep learning-based methods, including sequential modeling and graph-based baselines.
This highlights the merit of the end-to-end learning techniques for knowledge tracing.
However, traditional methods can achieve comparable or even better performance in some cases.
For example, KTM yields the second-best AUC score on EdNet.
Besides, one can see that although DKT has a very simple network structure, there is no constant winner between it and those more complex sequential modeling contenders, i.e., DKVMN, SAKT, and AKT.
Therefore, it is believed that more skillful techniques should be developed in modeling the learning sequences of students for knowledge tracing.

In Fig.~\ref{fig:AUC}, we also report the improvement of DGEKT over the best baseline in different categories.
Notably, while they all leverage the relationship information between exercises, DGEKT is substantially superior to the other two graph-based methods, i.e., GKT and PEBG, resulting in at least 2\% improvement on the three benchmark datasets.
The main reason may be attributed to the fact that in addition to exercise-concept associations, DGEKT also exploits the transitions between students' learning interactions, which are not fully explored in previous studies.
Moreover, it can be clearly seen that DGEKT exhibits a greater advantage on ASSIST17.
For example, the minimum improvement arises from 0.95\% and 0.28\% on ASSIST09 and EdNet to 6.84\% on ASSIST17, respectively.
We conjecture that this is because ASSIST17 contains more abundant exercise-answering records as listed in Table~\ref{tab:dataset}, from which DGEKT can establish a more dense directed transition graph of learning interactions and lead to better performance.

\begin{table}[t]
\caption{Setting differences between variant methods.}
\centering
\begin{tabular}{l c c c c c}
    \toprule
    \makecell[l]{Variant\\Method} & Hypergraph & \makecell[c]{Directed\\Graph} & CAHG & DTG & \makecell[c]{Knowledge\\Distillation} \\
    \midrule
    CAG & \ding{53} & \checkmark & \checkmark & \checkmark & \checkmark \\[2pt]
    TG & \checkmark & \ding{53} & \checkmark & \checkmark & \checkmark \\[2pt]
    RmCAHG  & \ding{53} & \checkmark & \ding{53} & \checkmark & \checkmark \\[2pt]
    RmDTG & \checkmark & \ding{53} & \checkmark & \ding{53} & \checkmark \\[2pt]
    RmOKD & \checkmark & \checkmark & \checkmark & \checkmark & \ding{53} \\[2pt]
    \bottomrule
\end{tabular}
\label{tab:variant_difference}
\end{table}

\begin{table}[t]
\caption{Performance comparison between DGEKT and its variant methods.}
\centering
\begin{tabular}{l c c c}
    \toprule
    Method & ASSIST09& ASSIST17 &EdNet \\
    \midrule
    CAG & 74.77\%  & 77.14\% & \underline{69.98\%} \\ [2pt]
    TG & 73.68\%  & \textbf{77.67\%} & 69.71\% \\ [2pt]
    RmCAHG & 73.02\%  & 76.00\% & 69.72\% \\ [2pt]
    RmDTG & 76.27\%  & 76.76\% & 69.28\% \\ [2pt]
    RmOKD & \underline{76.34\%}  & 76.89\% & 69.40\% \\ [2pt]
    DGEKT & \textbf{76.56\%}  &  \underline{77.54\%} & \textbf{70.07\%} \\ [2pt]
    \bottomrule
\end{tabular}
\label{tab:variant_performance}
\end{table}

\subsection{Ablation Study}
To further investigate the contributions of different components of DGEKT, we carried out a series of ablation studies by devising five variant methods of DGEKT:
\begin{itemize}
  \item \textbf{CAG} simplifies DGEKT via adopting a simple graph rather than a hypergraph in modeling the exercise-concept associations.
      In CAG, any pair of learning interactions are connected if they are associated with the same concept.
  \item \textbf{TG} simplifies DGEKT by using the undirected version of GCNs for modeling the transitions between learning interactions.
      It creates an undirected edge for each pair of consecutive interactions in students' learning sequences.
  \item \textbf{RmCAHG} removes the concept association hypergraph from DGEKT, meaning that only the transitions between learning interactions are leveraged to learn their embeddings.
  \item \textbf{RmDTG} removes the directed transition graph from DGEKT, meaning that only the exercise-concept associations are leveraged to learn the embeddings of learning interactions.
  \item \textbf{RmOKD} removes the procedure of online knowledge distillation in DGEKT.
        Instead, it concatenates the two representations of knowledge states generated by the dual graph models, and maps the result into the predicted probability via a sigmoid classifier layer.
\end{itemize}
Table~\ref{tab:variant_difference} summarizes the difference in the settings of these variant methods.
We show the performance comparison between them and DGEKT in Table~\ref{tab:variant_performance}, from which the following observations can be made:

\subsubsection{Rationality of Graph Type}
DGEKT models the exercise-concept associations and learning interaction transitions using hypergraphs and directed graphs, respectively.
When changing the graph types to simple undirected graphs, both CAG and TG experience performance degradation in most cases.
For example, CAG and TG are substantially inferior to DGEKT by 1.79\% and 2.88\% on ASSIST09, respectively.
The only exception is TG on ASSIST17, which is slightly better than DGEKT.
As aforementioned, this is perhaps because a more dense transition graph can be built on ASSIST17, leading to an ambiguous distinction between the directed and undirected graph settings.
The above observations confirm our belief that simply compressing the group-wise exercise-concept associations into pairwise ones could cause the loss of information.
At the same time, if the edge direction is ignored when diffusing the information along transition paths, it may deteriorate the performance.

\subsubsection{Contribution of Dual Graph Structure}
DGEKT integrates the dual graph structure consisting of CAHG and DTG to capture the heterogeneous relationships between learning interactions.
When removing either CAHG or DTG, RmCAHG and RmDTG are worse than DGEKT.
Besides, even without online knowledge distillation, RmOKD can usually obtain higher performance over RmCAHG and RmDTG by simply combining their outputs.
The results suggest that CAHG and DTG complement each other, and the dual graph structure is beneficial to improving knowledge tracing.

\begin{figure}[t]
    \centering
    \subfloat[DKT]
    {\includegraphics[width=0.3333\textwidth]{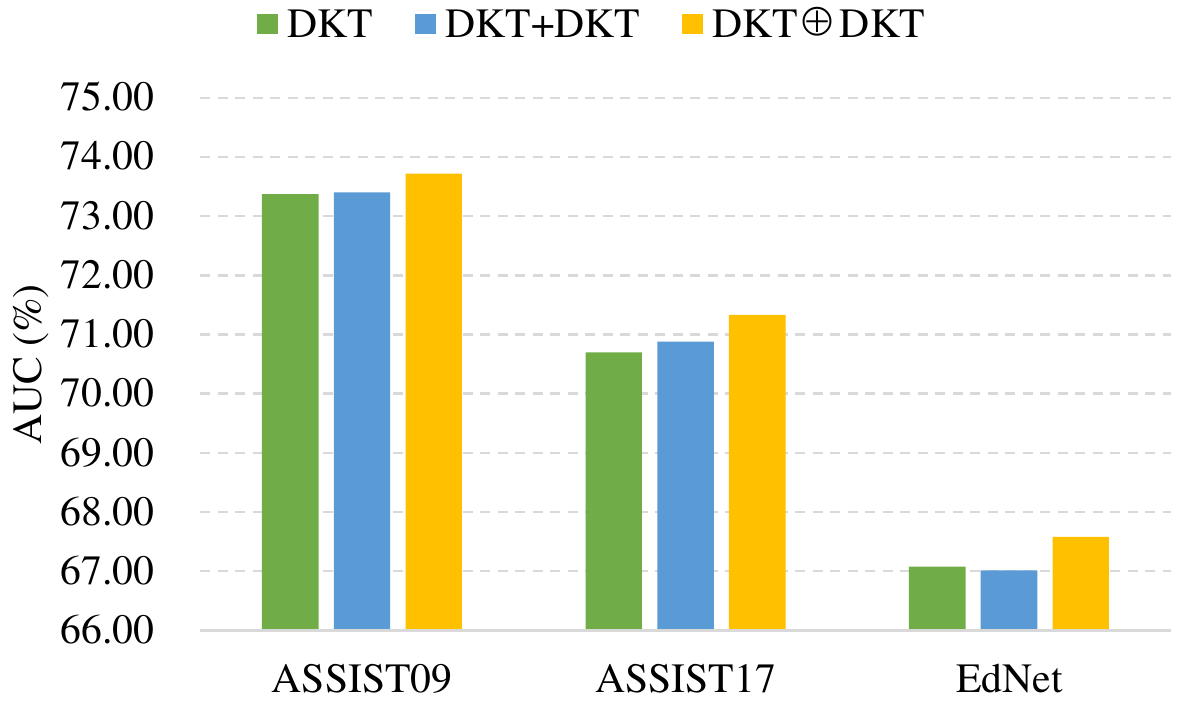}}
    \subfloat[SAKT]
    {\includegraphics[width=0.3333\textwidth]{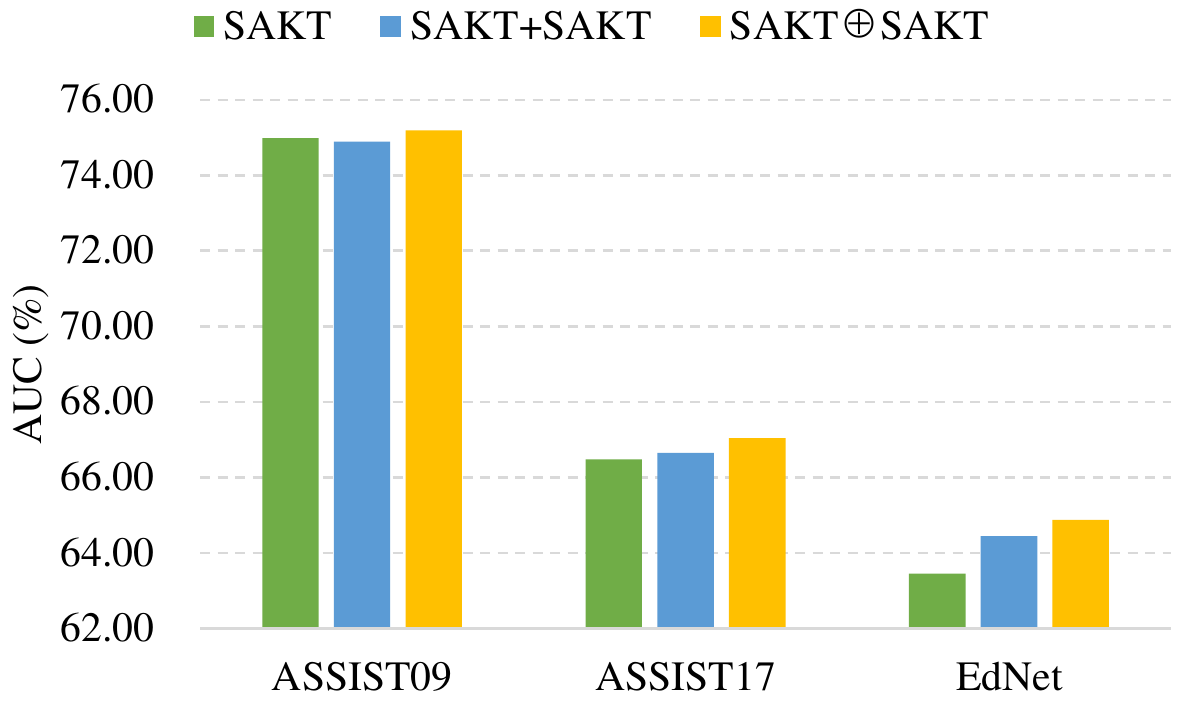}}
    \subfloat[DKT and SAKT]
    {\includegraphics[width=0.3333\textwidth]{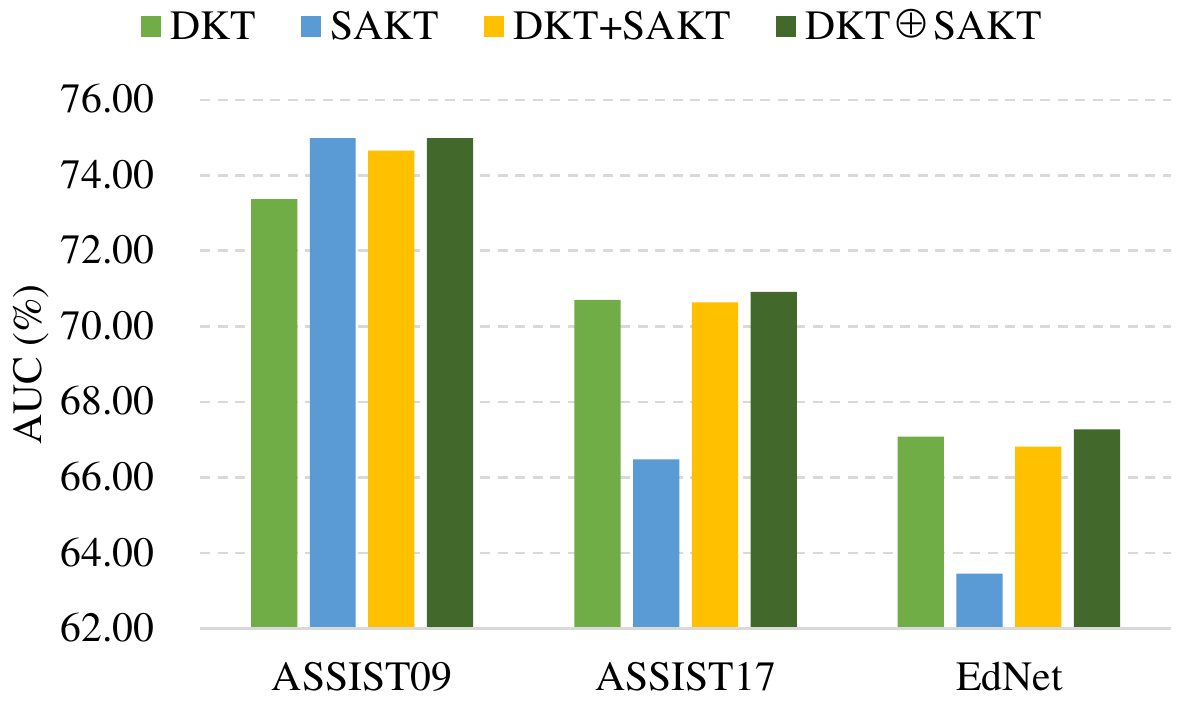}}
     \caption{Performance comparison between the concatenation scheme and online knowledge distillation for model ensemble in knowledge tracing.
     The symbols $+$ and $\oplus$ indicate the concatenation and online knowledge distillation based model ensemble, respectively.}
    \label{fig:OKD}
\end{figure}

\subsubsection{Benefit from Knowledge Distillation}
In DGEKT, we introduce the idea of online knowledge distillation to combine the dual graph models.
As demonstrated in Table~\ref{tab:variant_performance}, DGEKT consistently outperforms RmOKD on all datasets.
Also, we believe that online knowledge distillation can emerge as a highly effective solution for model ensemble in knowledge tracing.
To clarify this point, we separately adopt the concatenation scheme and online knowledge distillation to ensemble two DKT models, two SAKT models, as well as a DKT and a SAKT model.
The performance of the resulting ensemble models is displayed in Fig.~\ref{fig:OKD}, where the symbols $+$ and $\oplus$ indicate the concatenation and online knowledge distillation based ensemble, respectively.
Overall, the concatenation-based fusion may fail to produce extra performance gains, especially when ensembling the models with different architectures.
For example, DKT+SAKT obviously falls behind the better one of DKT and SAKT.
On the contrary, the strategy of online knowledge distillation offers a stable improvement over the single model, no matter through the homogeneous ensemble models (i.e., DKT$\oplus$DKT and SAKT$\oplus$SAKT) or the heterogeneous one (i.e., DKT$\oplus$SAKT).

\begin{figure}[t]
    \centering
    \subfloat[Graph embedding size]
    {\includegraphics[width=0.3333\textwidth]{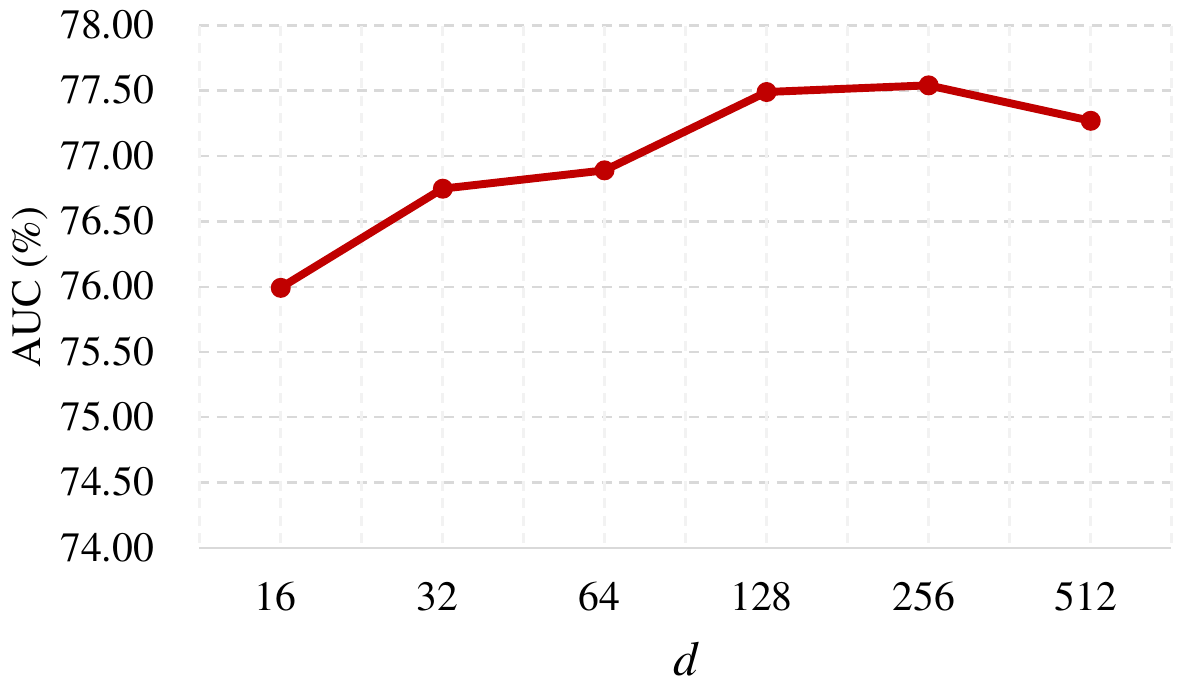}}
    \subfloat[Graph network depth]
    {\includegraphics[width=0.3333\textwidth]{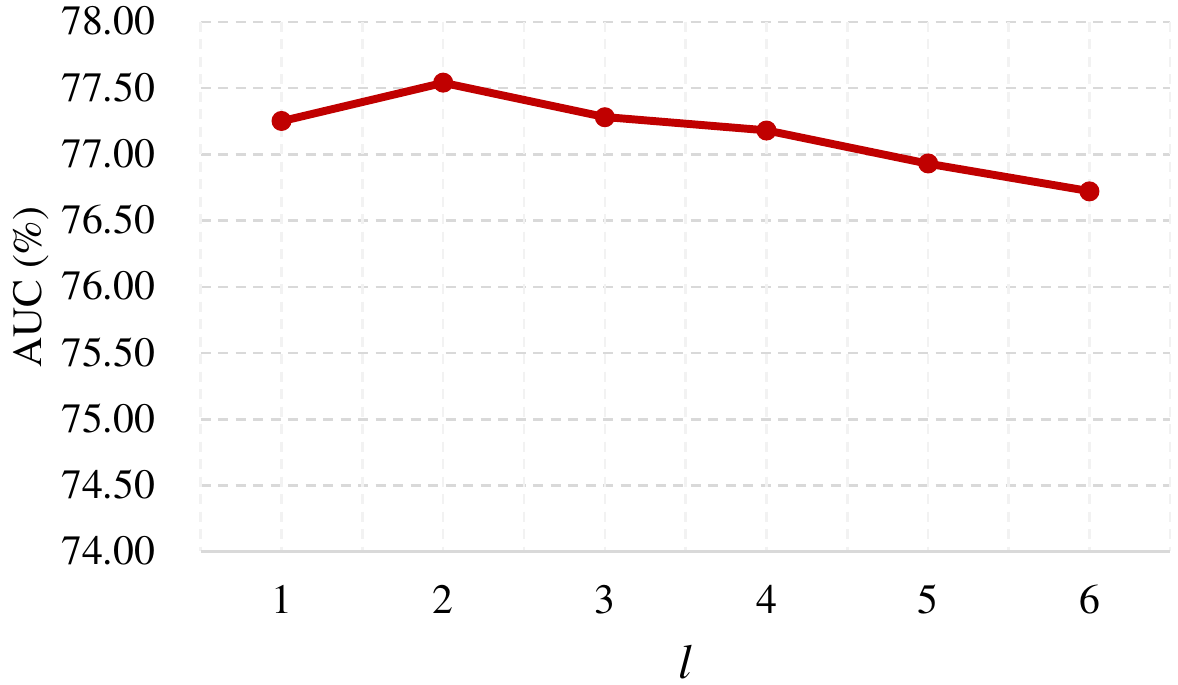}}
    \subfloat[Trade-off controller]
    {\includegraphics[width=0.3333\textwidth]{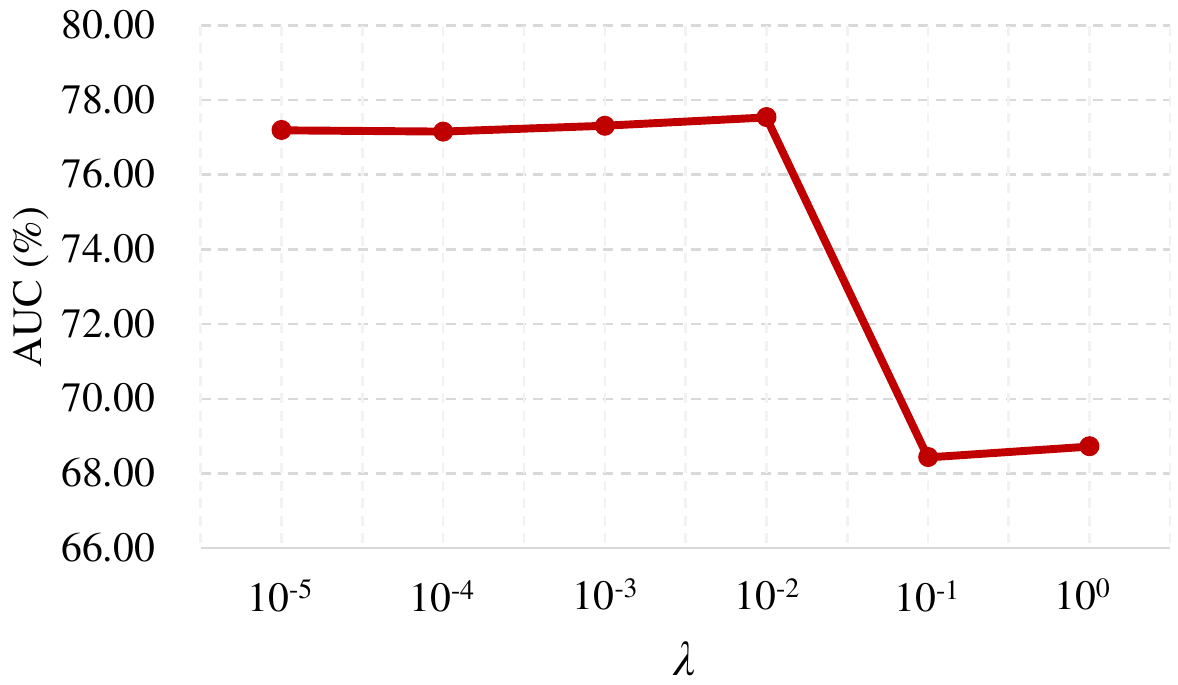}}
     \caption{Impact of different hyperparameters.}
    \label{fig:hyperparameter}
\end{figure}

\subsection{Sensitivity Analysis}
In order to investigate the sensitivity of DGEKT, we evaluate the impact of key hyperparameters on its performance.
Specifically, we consider the hyperparameters of embedding size $d$ and depth $l$ in the dual graph structure, as well as the trade-off controller $\lambda$ between the cross-entropy and distillation losses.
The experiments were carried out on ASSIST17, and the results are shown in Fig.~\ref{fig:hyperparameter}.

For the graph embedding size $d$, we impose the simplified assumption that different graph layers output the same size of node embeddings to reduce the complexity.
The value of $d$ is restricted to be a power of 2, ranging from 16 to 512.
As can be seen in Fig.~\ref{fig:hyperparameter}(a), the performance of DGEKT improves gradually with the increase of $d$, but overly increasing the embedding size could stagnate or even deteriorate the performance.
The best performance is achieved when $d=256$.
For the graph network depth $l$, we simply assume that CAHG and DTG have the same number of layers.
Fig.~\ref{fig:hyperparameter}(b) illustrates how the change of $l$ affects the performance of DGEKT.
We can observe that the performance goes down gradually when $l$ is larger than 2.
The phenomenon is mainly caused by the over-smoothing issue~\cite{li2018deeper} that the node representations become mixed and indistinguishable from each other after many layers of message passing in graph neural networks.
For the trade-off controller $\lambda$, we perform a logarithmic grid search from $10^{-5}$ to $10^0$ with a scaling factor of 10, as shown in Fig.~\ref{fig:hyperparameter}(c).
DGEKT keeps relatively steady performance when $\lambda$ is between $10^{-5}$ and $10^{-2}$, but rapidly declines with a further increase of $\lambda$.
On the whole, there is no drastic fluctuation in the performance of DGEKT when different hyperparameters vary within a wide range.
Therefore, it can be said that DGEKT has good model stability, and the optimal values of its hyperparameters are easily determined.

\subsection{Model Visualization}
\begin{figure}[t]
    \centering
    \subfloat[CAHG embeddings of learning interactions associated with ten knowledge concepts.]
    {\includegraphics[width=0.55\textwidth]{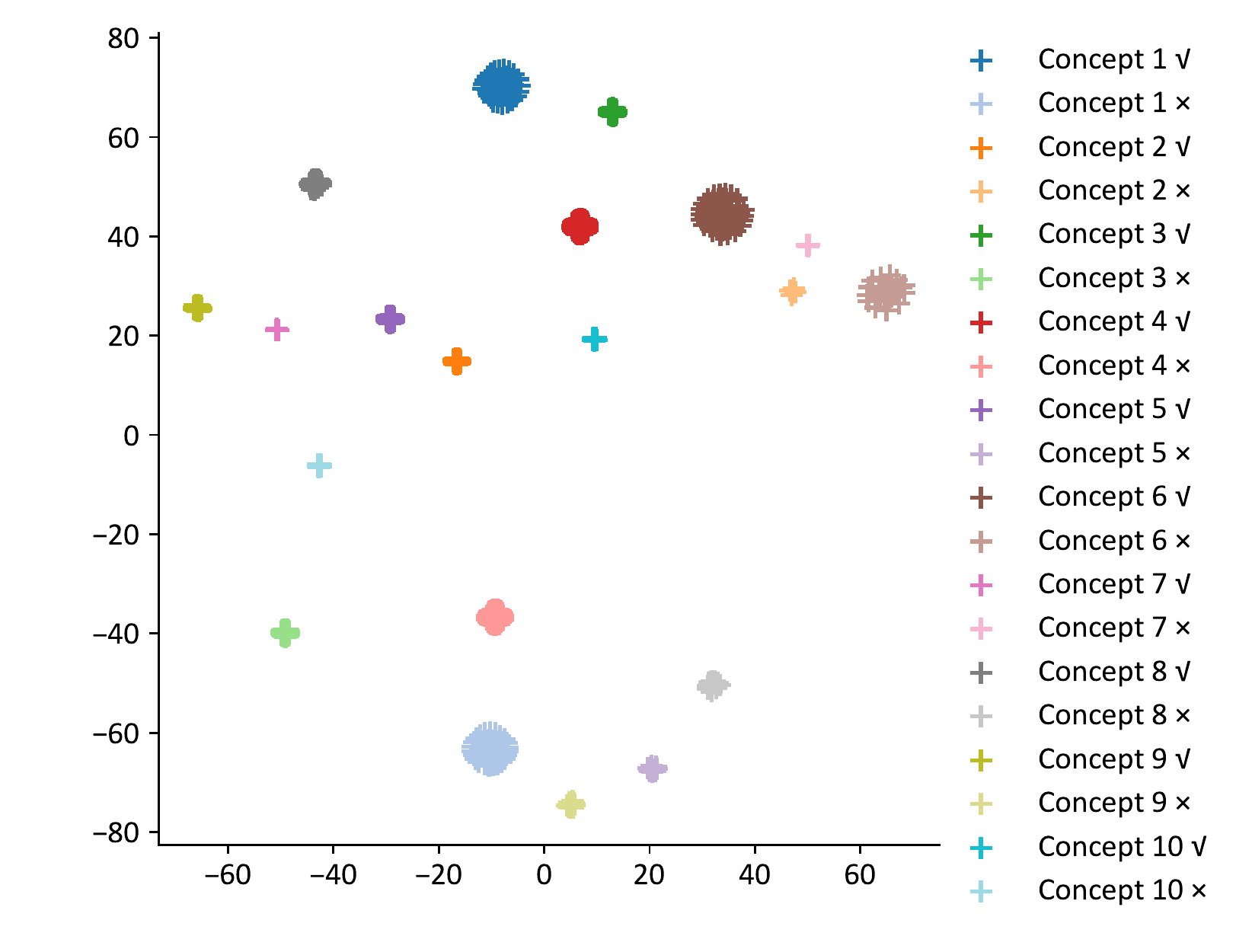}} \\
    \hspace{0.002\textwidth}
     \subfloat[CAHG embeddings of 50 randomly sampled knowledge concepts, among which five pairs of concepts with high co-occurrence frequency are indicated by different colored markers.]
    {\includegraphics[width=0.75\textwidth]{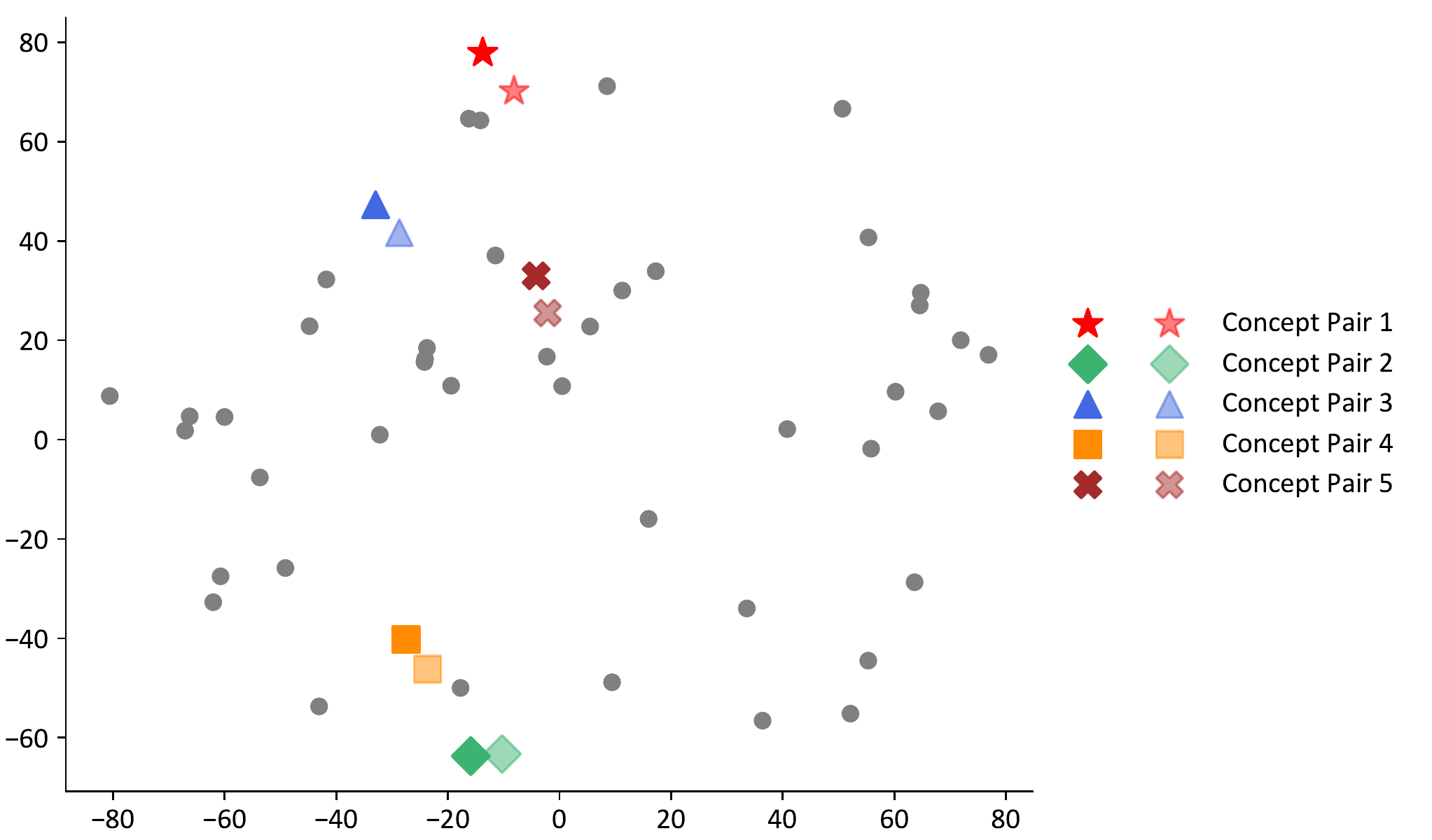}} \\
    \caption{Visualization of the learning interaction embeddings generated from CAHG.}
    \label{fig:visualize}
\end{figure}

In order to better understand the ability of DGEKT to represent students' learning interactions, we visualize the embeddings of learning interactions generated from CAHG and DTG.

Firstly, we randomly chose ten knowledge concepts, each of which has been decoupled into two hyperedges in CAHG, leading to a total of 20 hyperedges.
Then, the embeddings of learning interactions belonging to these hyperedges were projected into a two-dimensional space using the t-SNE algorithm~\cite{van2008visualizing}.
Fig.~\ref{fig:visualize}(a) displays the projection results.
As can be seen, learning interactions associated with different knowledge concepts can be well separated.
Meanwhile, we notice that the correct and incorrect responses to the same concept are usually distributed far away from each other.
The phenomenon is reasonable in the sense that correct and incorrect responses to the same concept somewhat reflect two opposite knowledge states, i.e., the mastery and non-mastery of the concept.
From these results, we can conclude that the information of exercise-concept associations is well integrated into our DGEKT model.

Besides, we find that DGEKT has the potential to discover the latent structure of knowledge concepts~\cite{piech2015deep}.
In our datasets, due to the lack of descriptions of knowledge concepts, the dependencies between concepts cannot be directly perceived.
Instead, we estimate the dependencies between two concepts by counting how many times they are associated with the same exercise.
Intuitively, if two concepts frequently appear together in the same exercise, they are likely to have the prerequisite or similarity relation in learning and education.
Fig.~\ref{fig:visualize}(b) shows the two-dimensional projections of 50 randomly sampled concepts.
Each concept is represented by the embedding center of learning interactions associated with the concept in CAHG.
In the figure, five pairs of concepts with high co-occurrence frequency are also indicated by different colored markers.
It can be clearly observed that the two concepts in each pair are embedded close to each other.
This suggests that we may infer the dependencies between knowledge concepts through the spatial distribution of their CAHG embeddings.

\begin{figure}
    \centering
    \includegraphics[width=0.8\textwidth]{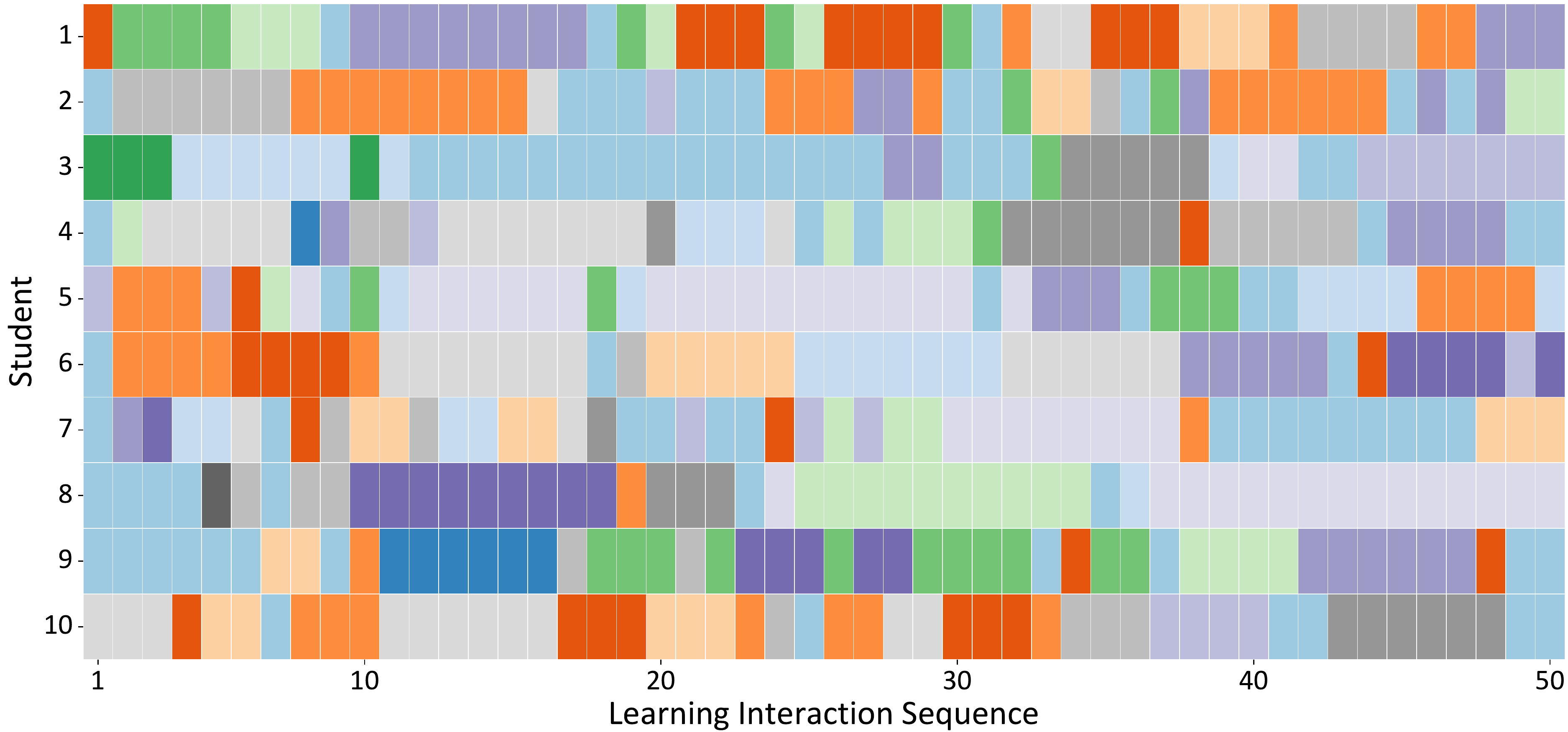}
    \caption{Visualization of the learning interaction sequences of ten students. All learning interactions are clustered into 20 groups based on DTG embeddings and distinguished by colors.}
    \label{fig:tEmbedding_vis}
\end{figure}

For DTG, we clustered all embeddings of learning interactions into 20 groups and randomly picked out ten students to exhibit their historical learning sequences.
In Fig.~\ref{fig:tEmbedding_vis}, the students' learning interactions are labeled by their respective clusters and distinguished by colors, and each row represents the learning sequence of a student.
It can be seen that consecutive interactions in a learning sequence are frequently partitioned into the same cluster, implying that they have approximate embeddings generated by DTG.
The finding demonstrates that DGEKT indeed captures the transitions between learning interactions across students with DTG.

\section{Conclusion}\label{section:conclusion}
In this paper, we investigate the dual graph structure of students' learning interactions that encodes the heterogeneous exercise-concept associations and interaction transitions.
In particular, we propose a dual graph ensemble learning method, named DGEKT, for knowledge tracing.
DGEKT integrates the two types of relationship information with hypergraph and directed graph convolutional networks.
To ensemble the dual graph models, we introduce the technique of online knowledge distillation rather than the straightforward concatenation- or adding-based fusion.
DGEKT leverages online knowledge distillation to form an ensemble teacher model from the dual graph models, which offers its predictions on all exercises as extra supervision for more powerful modeling capacity.
We performed a comparative study with eight previous methods on three benchmark datasets and verified the effectiveness of DGEKT for knowledge tracing.
We also demonstrated that the modules of dual graph learning and online knowledge distillation indeed benefit the performance of knowledge tracing.
A sensitivity analysis was performed to show that DGEKT has good model stability, and a visualization study was conducted to provide more insights into DGEKT.

Building upon the current study, our future work will be carried out in three directions.
Firstly, we intend to explore more elaborate network architectures instead of vanilla DKT to better model students' learning sequences.
In the experiments, we have noticed that the attention-based sequential modeling methods have no significant advantage over DKT.
It is an important issue to study how to make the attention mechanism play a more effective role in modeling learning sequences.
Besides, most exiting knowledge tracing methods only estimate the binary response (i.e., correctness or incorrectness) of students to exercises, but not all incorrect responses are equal: there can be numerous incorrect ways to answer an exercise caused by different underlying errors~\cite{Ghosh2021Option}.
Based on the recently released dataset~\cite{wang2021results} that contains the specific options students select on multiple choice exercises, we plan to extend our method beyond correctness prediction to the task of predicting students' option selections.
Finally, as pointed out in~\cite{long2022improving}, students may have close knowledge states if they behave similarly in learning.
Therefore, we will investigate how to measure the similarity between students' learning behaviors and explicitly leverage the inter-student information to enhance knowledge tracing.

\section*{Acknowledgments}
This work was partially supported by the National Natural Science Foundation of China (Grant No. 62077033 and No. 62177031),
the Shandong Provincial Natural Science Foundation Key Project (Grant No. ZR2020KF015),
the Shandong Provincial Natural Science Foundation (Grant No. ZR2021MF044)
and the Fostering Project of Dominant Discipline and Talent Team of Shandong Province Higher Education Institutions.

\bibliographystyle{unsrt}  
\bibliography{DGEKT_Ref} 

\end{document}